\documentclass{article} 
\usepackage{iclr2027_conference,times}

\usepackage{amsmath,amsfonts,bm}

\def\eqref#1{equation~\ref{#1}}

\def\1{\bm{1}}

\def\rvc{{\mathbf{c}}}

\def\rvs{{\mathbf{s}}}

\def\rmX{{\mathbf{X}}}

\DeclareMathAlphabet{\mathsfit}{\encodingdefault}{\sfdefault}{m}{sl}
\SetMathAlphabet{\mathsfit}{bold}{\encodingdefault}{\sfdefault}{bx}{n}

\usepackage{booktabs,tabularx,array}
\usepackage{hyperref}
\usepackage{url}
\usepackage{amsmath}
\usepackage{amssymb}
\usepackage{booktabs}
\usepackage{bm}
\usepackage{multirow}
\usepackage{threeparttable}
\usepackage{graphicx}
\usepackage{listings}
\usepackage{microtype}
\usepackage[most]{tcolorbox}
\usepackage{siunitx}
\usepackage{float}
\usepackage{xurl}
\providecommand{\dscheck}{\ensuremath{\checkmark}}

\providecommand{\dscheck}{\ensuremath{\checkmark}}
\tcbuselibrary{listings}

\newcommand{\myfnsymbol}[1]{%
  \ifcase#1\or *\or \dagger\or \ddagger\or \S\or \P\fi
}

\newcommand{\tabfn}[1][1]{%
  \textsuperscript{\ensuremath{\myfnsymbol{#1}}}%
}
\title{AVIO: Learning to Add and Remove Sounding Objects in Audiovisual Scenes}

\author{%
\begin{minipage}{0.96\textwidth}
\raggedright
\vspace*{1.5ex}
\mbox{\textbf{Weihan Xu}\textsuperscript{1, 2}} \quad
\mbox{\textbf{Kan Jen Cheng}\textsuperscript{3}} \quad
\mbox{\textbf{Koichi Saito}\textsuperscript{5}} \quad
\mbox{\textbf{Jingyu Shi}\textsuperscript{6}} \\[0.5ex]
\mbox{\textbf{Tingle Li}\textsuperscript{4}} \quad
\mbox{\textbf{Yisi Liu}\textsuperscript{4}} \quad
\mbox{\textbf{Liming Wang}\textsuperscript{2}} \quad
\mbox{\textbf{Masato Ishii}\textsuperscript{5}} \quad
\mbox{\textbf{Takashi Shibuya}\textsuperscript{5}} \\[0.5ex]
\mbox{\textbf{Gopala Anumanchipalli}\textsuperscript{4}} \quad
\mbox{\textbf{Paul Pu Liang}\textsuperscript{2}}\\[1.5ex]
\normalfont\small
\textsuperscript{1}University of Washington \quad
\textsuperscript{2}Massachusetts Institute of Technology\\[0.3ex]
\textsuperscript{3}University of Maryland, College Park \quad
\textsuperscript{4}University of California, Berkeley \\[0.3ex]
\textsuperscript{5}Sony Research \quad
\textsuperscript{6}Independent Researcher \quad
\\[0.7ex]
\footnotesize\href{mailto:wx35@uw.edu}{\texttt{wx35@uw.edu}}
\end{minipage}%
}

\newcommand{\mypar}[1]{{\bf #1} }

\iclrfinalcopy
\begin{document}

\maketitle

\begin{abstract}
Adding or removing a sounding object requires coordinated changes to
visual content and sound while preserving the surrounding scene.
Yet paired supervision for localized non-speech audiovisual editing
remains limited, as visual and acoustic edits must target the same
object and isolate its sound from overlapping sources.
To address this gap, we introduce \textit{AVIOBench},
a dataset comprising 37.9 hours of paired audiovisual examples
spanning 1{,}878 target-object names.
AVIOBench links the visual presence and acoustic contribution of
each target object through a shared identity and visual mask.
Our automated pipeline
uses visual grounding and cross-modal consistency to select
target-sound removal candidates, then jointly refines the audiovisual
pairs to improve perceptual quality and cross-modal consistency.
Building on this dataset, we propose \textit{AVIO}, which adapts a pretrained
text-to-audiovisual generation model through source-conditioned
feature modulation to jointly learn object addition and removal.
A reference-frame curriculum gradually reduces reference conditioning
during training, enabling one model to perform instruction-only
editing with optional visual guidance.
Quantitative and qualitative evaluations demonstrate effective
audiovisual object removal and addition, with optional reference
guidance providing appearance and placement control for addition.
Video examples are available on our project
page\footnote{\url{https://wx83.github.io/AVIO-Object-Level-Addition-and-Removal-for-Audiovisual-Scenes/}}.
\end{abstract}

\section{Introduction}
\begin{figure*}[t]
    \centering
\includegraphics[
    width=\textwidth,
    height=0.4\textheight,
    keepaspectratio
]{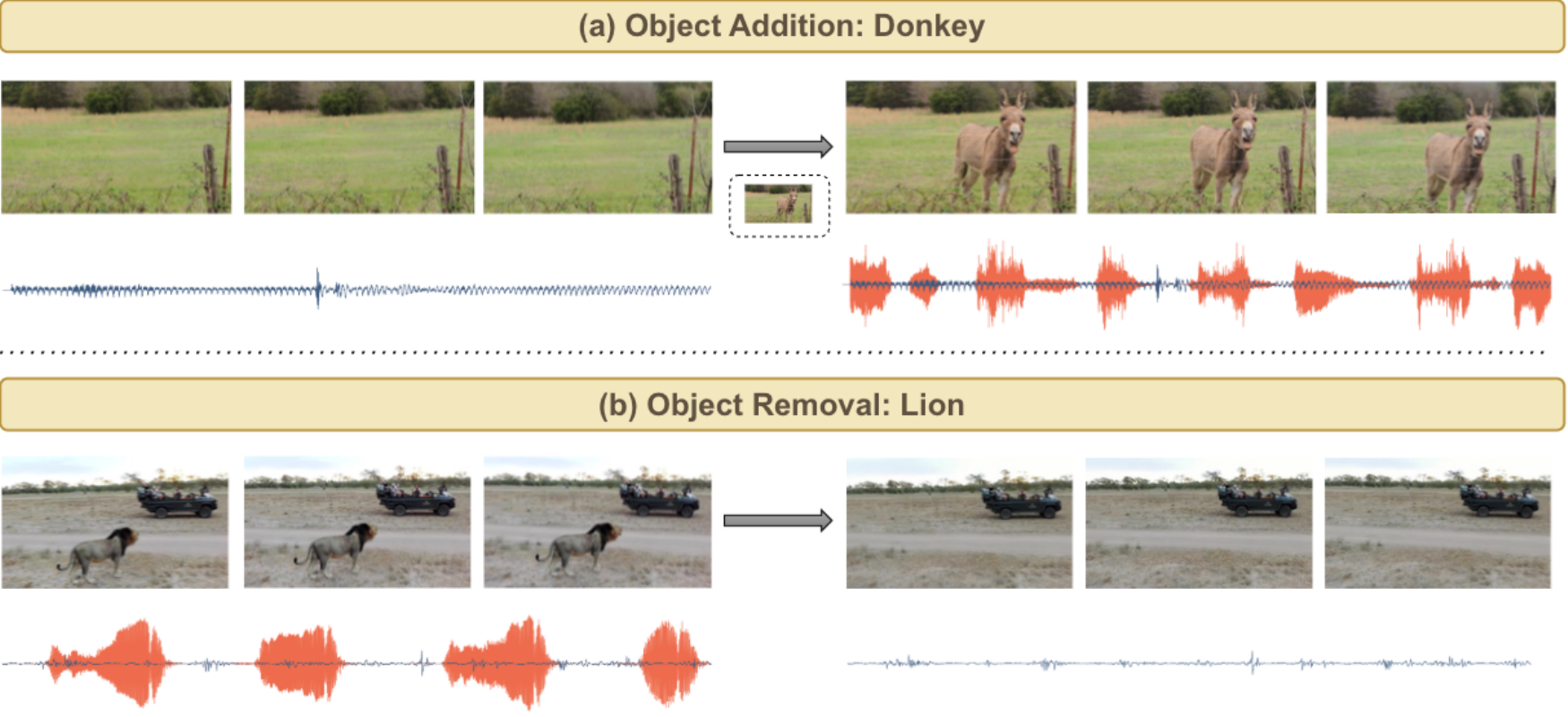}
    \vspace{-2ex}
    \caption{
        \textbf{Audiovisual event addition and removal with AVIO.}
        (a) A donkey and its braying are added to the scene with an optional reference frame to specify initial placement and appearance
        (b) A lion and its roaring are removed while the vehicle and its sound are preserved.
        }
    \label{fig:teaser}
\end{figure*}

Adding or removing a sounding object requires
editing both its visual presence and its associated sound.
As shown in Fig.~\ref{fig:teaser}, removing a donkey should also eliminate its
braying, while adding a passing vehicle should introduce
sound consistent with its motion and timing.
Such joint audiovisual editing supports acoustically coherent
film post-production and content creation~\citep{fu2025objectavedit},
and could complement robotic and driving
simulators~\citep{chen2022soundspaces2,dosovitskiy2017carla}
by jointly varying objects and sounds to test audiovisual
perception robustness.
However, unimodal editors cannot jointly edit visual objects and their associated sounds, while audiovisual editors and benchmarks lack a focus on sounding objects, partly due to limited paired supervision~\citep{lin2025zeroshotaudiovisualeditingcrossmodal,fu2025objectavedit,zheng2026aviedit,zheng2026instructav2av,chen2026javeditjointaudiovisualinstructionguided,wen2026avecompassholisticevaluationaudiovideo,miao2026omnieditbenchcomprehensivebenchmarkinstructionbased}.

Constructing paired supervision for this setting poses two
challenges: isolating non-speech sounds from complex mixtures
and coordinating object-level changes across audio and video
while preserving unrelated content.
Environmental sounds exhibit diverse spectral and temporal
patterns~\citep{chu2009environmental} and often overlap with
other sources, making the target sound difficult to isolate
without altering unrelated sounds.
Beyond acoustic separation, the visual and acoustic edits
must refer to the same object and consistently change its
presence in both modalities.

To address this gap, we introduce \textit{AVIOBench}, comprising
approximately \textbf{37.9 hours}\footnote{The duration of each source--target pair is counted once.} of paired audiovisual examples
constructed from AudioSetCaps and
VGGSound~\citep{bai2024audiosetcaps,chen2020vggsoundlargescaleaudiovisualdataset},
covering diverse object-associated sound events beyond
speech.
In order to remove the same sounding object across both
modalities while preserving unrelated content, our pipeline
first identifies the target object and uses it to guide
both visual removal and audio removal.
To address the difficulty of removing sound, we generate separation candidates using varied text prompts, visual-mask prompts, and random seeds, then select the candidate with the strongest audiovisual correspondence to the target object.
In order to further improve consistency between the resulting visual
and acoustic edits, we apply joint audiovisual refinement on the edited pairs.  

Building on AVIOBench, we introduce \textit{AVIO}, an object-level audiovisual editor that adapts a pretrained text-to-audiovisual
generation model, drawing on its learned modality representations and its
temporally aligned joint audio–video architecture. To enable source-consistent editing, we introduce source-conditioned feature modulation that aligns source features with target representations in time and modality. While text instructions specify the desired edit, they may leave the appearance and initial placement of the added object ambiguous. To reduce this ambiguity, AVIO optionally uses an edited
reference frame to guide appearance and initial placement.
A reference-frame curriculum progressively reduces reference
availability during training, enabling editing with text alone
or text together with a reference frame. Through quantitative evaluation and qualitative comparisons, we demonstrate the effectiveness of AVIO in audiovisual object removal and addition, with optional reference guidance providing appearance and placement control for addition.


Our contributions are summarized as follows:

\begin{itemize}
    \item We introduce AVIOBench, a dataset of \textbf{37.9 hours}
    of paired audiovisual examples for object addition and
    removal involving non-speech sound events.
    \item We adapt a pretrained text-to-audiovisual Diffusion Transformer (DiT) into
    a unified object editor jointly trained for addition and
    removal.
    A reference-frame curriculum supports both text-only
    editing and optional visual guidance specifying the appearance and initial placement of the target object.
\end{itemize}

We will publicly release AVIOBench, our code, and a LangGraph-based data engine to support reproducibility and further dataset expansion, subject to applicable licenses upon acceptance.

\section{Related Work}

\mypar{Object Editing Datasets.} 
Existing editing datasets differ in whether they supervise changes within a single modality or coordinate edits across audio and video. Visual datasets such as OpenVE-3M and Goku~\citep{he2026openve3mlargescalehighqualitydataset,liang2026gokumillionscaleuniversaldataset}, and audio resources such as AUDIT~\mbox{~\citep{wang2023auditaudioeditingfollowing}}, provide single-modality supervision. Audiovisual resources extend this setting to object editing~\mbox{\citep{lin2025zeroshotaudiovisualeditingcrossmodal,fu2025objectavedit}}, paired instruction-guided editing~\citep{chen2026javeditjointaudiovisualinstructionguided,zheng2026instructav2av}, and broader cross-modal evaluation~\citep{wen2026avecompassholisticevaluationaudiovideo,miao2026omnieditbenchcomprehensivebenchmarkinstructionbased}. AVIOBench focuses on non-speech sounding objects, providing aligned object-present and object-absent audiovisual pairs that support both addition and removal. 
A detailed dataset comparison can be found in Appendix ~\ref{sec:dataset_comp}.

\mypar{Audiovisual Joint Generation.} 
Earlier audiovisual methods generate one modality conditioned
on the other, while recent work shifts toward joint generation.
MM-Diffusion~\citep{ruan2022mmdiffusion} couples audio and video
diffusion networks, while JavisDiT and
JavisDiT++~\citep{liu2025javisdit,liu2026javisditunifiedmodelingoptimization}
use Transformers to model cross-modal interactions.
Extending generation to editing requires
coordinated changes to a target object and its sound while
preserving unrelated source content.
To enable such editing, we adapt a pretrained audiovisual
generator through token-aligned source modulation and paired
editing supervision to jointly learn object addition and removal.

\mypar{Audiovisual Joint Editing.}
Existing audiovisual editing methods coordinate visual and
acoustic changes through sequential cross-modal guidance
or joint audiovisual modeling.
AvED~\citep{lin2025zeroshotaudiovisualeditingcrossmodal} and
Object-AVEdit~\citep{fu2025objectavedit} use delta denoising
and inversion-based regeneration, respectively.
CoherentAVEdit~\citep{ishii2026coherentaudiovisualeditingconditional}
generates audio after video editing, whereas
AVI-Edit~\citep{zheng2026aviedit} uses audio to guide visual edits.
SpongeBob~\citep{liang2026spongebobsyncawareharmoniousaudiovisual}
couples modalities during denoising, while
JAVEdit~\citep{chen2026javeditjointaudiovisualinstructionguided}
and InstructAV2AV~\citep{zheng2026instructav2av}
adapt audiovisual generators using paired editing data.
We focus on non-speech object addition and removal,
learning bidirectionally from object-present and object-absent
pairs with a reference-frame curriculum supporting text-only
editing and optional visual guidance for addition.

\begin{figure*}[t]
  \centering
  \includegraphics[width=\linewidth]{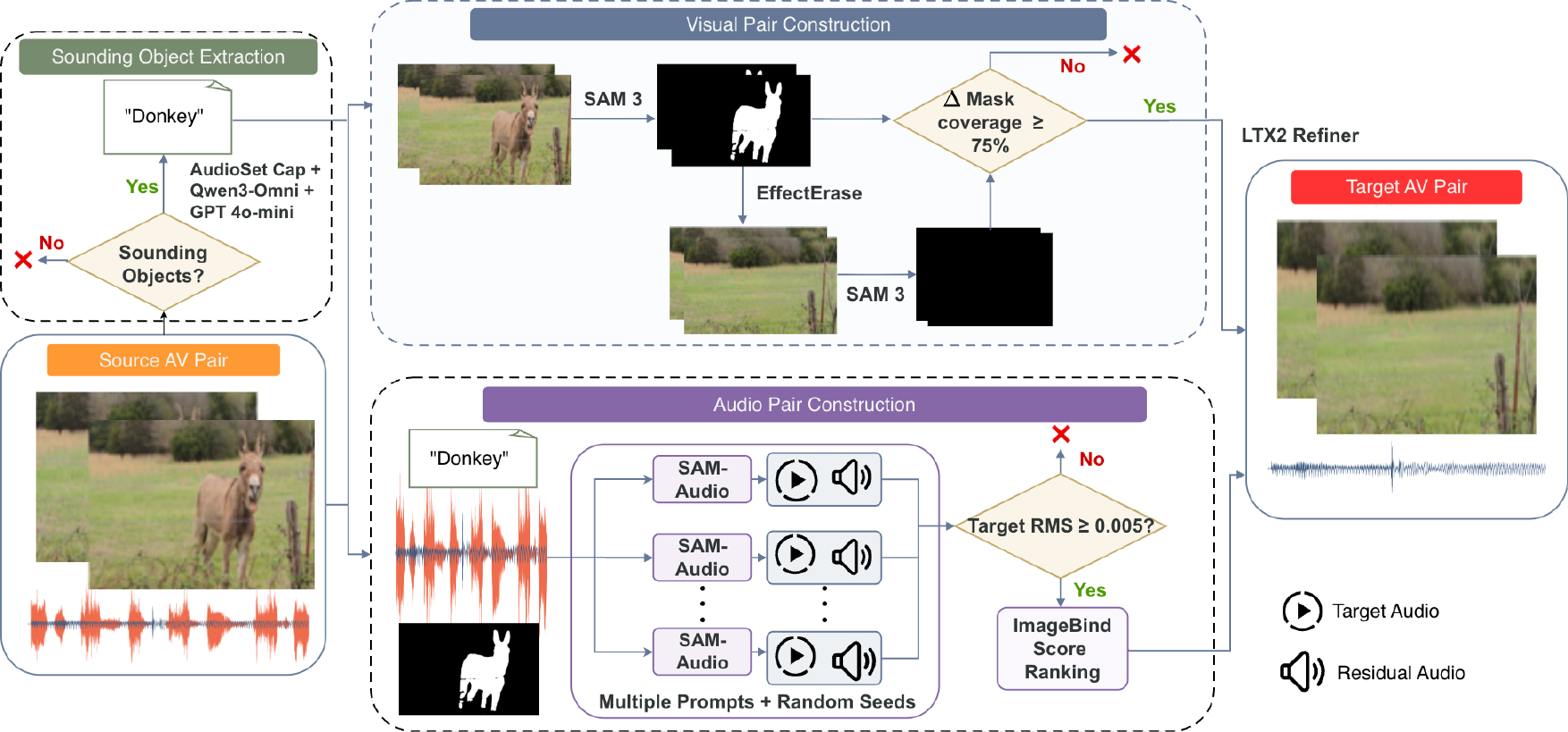}\vspace{-2ex}
\caption{\textbf{Overview of the AVIOBench data construction pipeline.} For each clip, we identify the primary sounding object $O$, remove it visually via segmentation and inpainting, and remove its sound via audio separation. Each stage applies automatic filters (yellow) to discard unreliable candidates, and the edited streams are jointly refined to form $\mathbf{X}_{\setminus O}$.}
  \label{fig:data_pipeline}
  \vspace{-1mm}
\end{figure*}

\section{AVIOBench Dataset}

We introduce AVIOBench, a benchmark for joint audiovisual
editing of object-associated sound events, with paired examples
for object removal and addition.
We describe the dataset construction pipeline shown in
Fig.~\ref{fig:data_pipeline} in
Section~\ref{sec:dataset_construction_pipeline}
and summarize the dataset size and annotations in
Section~\ref{sec:dataset_statistics}.
\subsection{Data Construction Pipeline}\label{sec:dataset_construction_pipeline}

\mypar{Data Collection and Filtering.}
We collect source videos from AudioSet~\citep{gemmeke2017audioset} and VGGSound~\citep{chen2020vggsoundlargescaleaudiovisualdataset}, two large-scale datasets of 10-second YouTube clips captured in diverse real-world settings. These datasets span human activities, animals, transportation, household sounds, musical performances, and natural environments. For computational efficiency, we retain the first five seconds of each clip, processing 159 hours of video in total. To focus on object-associated sound events, we use FireRedASR2S~\citep{xu2026fireredasr2sstateoftheartindustrialgradeallinone} to exclude clips containing only speech.

\mypar{Target-Object Identification.}
To identify a primary sounding object in each clip
$\mathbf{X}=(V,A)$, we first obtain an audiovisual caption.
We use the caption from AudioSetCaps~\citep{bai2024audiosetcaps}
when available and otherwise generate one using
Qwen3-Omni~\citep{xu2025qwen3omnitechnicalreport}.
Given this caption, GPT-4o mini~\citep{openai2024gpt4omini}
identifies the primary sounding object $O$, which serves as
the prompt for \textsc{SAM~3}~\citep{carion2026sam3segmentconcepts}
to extract its video mask $M_O$.

\mypar{Audio Pair Construction.}
We construct an audio pair from the original recording and a residual
track with the target sound removed.
Specifically, we run \textsc{SAM-Audio}~\citep{shi2025samaudiosegmentaudio}
with the object name and visual mask as prompts across multiple random
seeds, jointly generating target and residual stems.
We discard candidates whose target stem has a root-mean-square (RMS)
amplitude below 0.005 to exclude near-silent extractions.
Among the remaining candidates, we select the target stem $a_O$ with
the highest ImageBind~\citep{girdhar2023imagebindembeddingspacebind}
similarity to the mask-conditioned video and use its corresponding
residual as the object-absent audio.

\mypar{Visual Pair Construction.}
We construct the corresponding visual pair by removing the target object through video inpainting. We apply \textsc{EffectErase}~\citep{fu2026EffectErase} to the region specified by $M_O$, producing the edited video $V_{\setminus O}$. To assess whether the object remains detectable, we rerun \textsc{SAM~3}~\citep{carion2026sam3segmentconcepts} with the same object prompt. We retain only clips for which the mean detected mask area decreases by at least $75\%$ relative to the original video. 

\mypar{Audiovisual Pair Construction.}
We combine the edited streams into
$\mathbf{X}_{\setminus O}=(V_{\setminus O},A_{\setminus O})$. 
Following Ditto~\citep{bai2025scalinginstructionbasedvideoediting},
we apply a lightweight SDEdit refinement with an empty prompt,
using LTX-2~\citep{hacohen2026ltx2efficientjointaudiovisual}
to jointly refine both modalities and reduce editing artifacts.
We use $\mathbf{X}_{\setminus O}$ to denote the resulting refined clip
and construct removal and addition examples by exchanging the source
and target:
\begin{equation}
(\mathbf{X}_{\mathrm{s}},\mathbf{X}_{\mathrm{e}})
=
\begin{cases}
(\mathbf{X},\mathbf{X}_{\setminus O}), & \text{removal},\\
(\mathbf{X}_{\setminus O},\mathbf{X}), & \text{addition}.
\end{cases}
\end{equation}

\subsection{Dataset Statistics}
\label{sec:dataset_statistics}

AVIOBench contains 37.9 hours of unique paired examples with object-level
text annotations and masks. Each example is
annotated with an object name (1{,}878 unique names in total)
and a mask of the target object. We provide additional dataset statistics and retention rates
at each pipeline stage and the details of our LangGraph-based data engine in Appendix~\ref{sec:dataset_details}.


\vspace{-2ex}

\section{AVIO Model}
\label{sec:clickav2av}
\begin{figure*}
  \centering
  \includegraphics[width=1\linewidth]{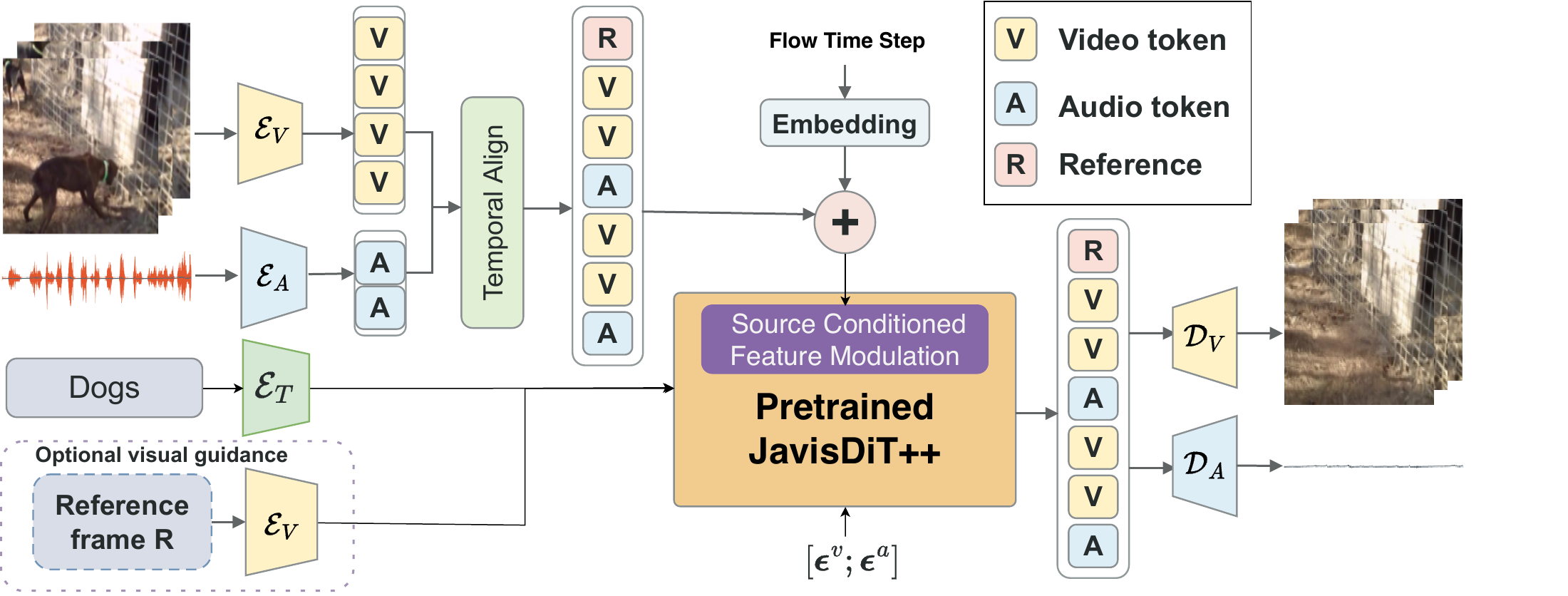}\vspace{-2ex}
    \caption{\textbf{Overview of AVIO.}We adapt a pretrained text-to-audiovisual DiT by conditioning it on source audiovisual latents. To support addition and removal within a unified model, we introduce a reference-guided curriculum that progressively increases the dropout probability of the first target frame during training.}
  \label{fig:architecture}
  \vspace{-1ex}
\end{figure*}

We introduce AVIO, which adapts a pretrained text-to-audiovisual
generation model for joint object addition and removal.
We retain its text-conditioning branch and add source-conditioned
feature modulation to enable instruction-guided editing
(Section~\ref{sec:clickav2av_backbone}).
We then present a reference-frame curriculum for optional
visual guidance (Section~\ref{sec:clickav2av_curriculum}),
followed by training and inference
(Section~\ref{sec:clickav2av_inference}).

\mypar{Problem formulation.} Let $\mathcal{M}=\{v,a\}$ denote the video and audio modalities. 
Given a source audiovisual pair $\rmX_{\mathrm{s}}=(\rmX_{\mathrm{s}}^v,\rmX_{\mathrm{s}}^a)$ and editing conditions $\rvc=\{I,R\}$, where $I$ is a textual instruction and $R$ is an optional edited reference frame, our goal is to predict the edited target $\rmX_{\mathrm{e}}=(\rmX_{\mathrm{e}}^v,\rmX_{\mathrm{e}}^a)$.


\subsection{Instruction-Guided Audiovisual Editing}
\label{sec:clickav2av_backbone}

\mypar{Pretrained Backbone and text conditioning.}
In order to adapt pretrained audiovisual generation to
object-level editing, we build on the learned modality
representations and temporally aligned joint audio--video
architecture of
JavisDiT++~\citep{liu2026javisditunifiedmodelingoptimization}.
We retain its text-conditioning branch to guide the desired edit.
The pretrained umT5-XXL encoder
$\mathcal{E}_T$~\citep{chung2023unimax} maps the instruction
$I$ to features $\mathcal{E}_T(I)$, which are supplied to
the text cross-attention layers of the DiT blocks.

\mypar{Source encoding and alignment.}
To preserve source content unrelated to the requested edit,
AVIO aligns source features with target tokens and uses
these features to guide editing through feature-wise linear
modulation (FiLM)~\citep{perez2018film}.
As shown in Fig. ~\ref{fig:architecture}, we encode and tokenize the source and target following
JavisDiT++~\citep{liu2026javisditunifiedmodelingoptimization}
to maintain their spatial and temporal alignment.
Specifically, let $\mathcal{E}_V$ and $\mathcal{E}_A$ denote
the frozen Wan2.1 video VAE encoder~\citep{wan2025} and
AudioLDM2 audio VAE encoder~\citep{liu2024audioldm2learningholistic},
respectively, with audio preprocessing included in $\mathcal{E}_A$.
For modality $m\in\{v,a\}$, denoting video and audio,
we encode the source input $\rmX_{\mathrm{s}}^m$ into a latent
tensor $\mathbf{z}_{\mathrm{s}}^m$.
We then apply learnable patch embeddings $\psi_v$ and $\psi_a$
to the video and audio latents, respectively, and concatenate
the resulting token sequences into the source representation
$\rvs$:
\begin{equation}
\begin{aligned}
    \mathbf{z}_{\mathrm{s}}^m
    &= \mathcal{E}_m(\rmX_{\mathrm{s}}^m),
    \qquad m\in\{v,a\},\\
    \rvs
    &= \left[
        \psi_v(\mathbf{z}_{\mathrm{s}}^v);
        \psi_a(\mathbf{z}_{\mathrm{s}}^a)
    \right],
\end{aligned}
\label{eq:source_tokens}
\end{equation}
where $[\cdot;\cdot]$ denotes concatenation along the token
dimension.
For cross-modal temporal alignment, we retain temporal-aligned
rotary positional embeddings (TA-RoPE) in the backbone's joint
self-attention, assigning a shared temporal position index
to audio and video tokens corresponding to the same temporal
interval.

\mypar{Source-conditioned feature modulation.}
To promote source--target alignment, we modulate hidden
states of the intermediate flow sample $\mathbf{z}_{\tau}$
with source features through adaptive layer normalization
(AdaLN)~\citep{peebles2023dit}, following
FiLM~\citep{perez2018film}.
For hidden state $\mathbf{x}_i$ at position $i$ of the joint
audiovisual sequence, we combine the corresponding source
token $\mathbf{s}_i$ (Eq.~\ref{eq:source_tokens}) with the
flow-time embedding $\mathbf{e}_{\tau}$:
\begin{equation}
    \mathbf{h}_i = \mathbf{e}_{\tau} + \mathbf{s}_i.
    \label{eq:source_cond_vec}
\end{equation}
Each self-attention ($\mathrm{sa}$), text cross-attention
($\mathrm{ca}$), and modality-specific feed-forward
($\mathrm{ff}$) sublayer $k$ applies
\begin{equation}
    \begin{aligned}
        \left(\boldsymbol{\beta}_i^{k},
        \boldsymbol{\gamma}_i^{k},
        \boldsymbol{\alpha}_i^{k}\right)
        &= \mathcal{M}^{k}(\mathbf{h}_i) + \boldsymbol{\delta}^{k}, \\
        \widetilde{\mathbf{x}}_i
        &= (\mathbf{1}+\boldsymbol{\gamma}_i^{k})
        \odot \operatorname{LN}(\mathbf{x}_i)
        + \boldsymbol{\beta}_i^{k}, \\
        \mathbf{x}_i^{\mathrm{out}}
        &= \mathbf{x}_i + \boldsymbol{\alpha}_i^{k}
        \odot [\mathcal{F}^{k}(\widetilde{\mathbf{X}})]_i,
    \end{aligned}
    \label{eq:source_adaln}
\end{equation}
where $\mathcal{M}^{k}$ is a SiLU--linear head predicting
shift $\boldsymbol{\beta}_i^{k}$, scale adjustment
$\boldsymbol{\gamma}_i^{k}$, and residual gate
$\boldsymbol{\alpha}_i^{k}$.
We reuse the pretrained heads and learned per-block offsets
$\boldsymbol{\delta}^{k}$ for self-attention and feed-forward
sublayers, with modality-specific feed-forward heads.
For the backbone's unmodulated text cross-attention, we add
a head $\mathcal{M}^{\mathrm{ca}}$ shared across blocks with
$\boldsymbol{\delta}^{\mathrm{ca}}=\mathbf{0}$.
$\mathcal{F}^{k}$ operates on the modulated sequence
$\widetilde{\mathbf{X}}=[\widetilde{\mathbf{x}}_1,\ldots,
\widetilde{\mathbf{x}}_L]$, with text cross-attention using
the encoded instruction $\mathcal{E}_T(I)$ as keys and values.
$\operatorname{LN}$ denotes layer normalization,
$\odot$ element-wise multiplication, and $[\cdot]_i$ the
output at position $i$. 
Block indices and hidden-state sublayer indices are omitted for clarity.

\subsection{Reference-Guided Curriculum}
\label{sec:clickav2av_curriculum}
In order to support reference-guided addition and
appearance-controlled editing alongside removal without
a reference frame, we adopt a reference-guided curriculum
inspired by Ditto~\citep{bai2025scalinginstructionbasedvideoediting}.
The curriculum initially provides visual guidance and
gradually reduces its availability, allowing a single model
to learn editing under both conditioning regimes. We validate this design in our curriculum ablation in Appendix ~\ref{sec:ablation_curriculum}. 
To specify the desired object's appearance and initial
placement, we provide an optional edited reference frame $R$.
To incorporate this visual guidance through the existing
source pathway, we encode $R$ into a single latent frame
$\mathbf{z}_R=\mathcal{E}_V(R)$ and prepend it to the
source video latent:
\begin{equation}
    \tilde{\mathbf{z}}_{\mathrm{s}}^v
    =
    \left[
        \mathbf{z}_R;\mathbf{z}_{\mathrm{s}}^v
    \right],
    \label{eq:reference_prefix}
\end{equation}
where concatenation is along the temporal dimension.
In order to preserve source--target token correspondence, we allocate
a matching temporal prefix slot in the target video latent
and discard its output at inference as shown in Fig. ~\ref{fig:architecture}.  
During training, we use the first frame of the target video
$\rmX_{\mathrm{e}}^v$ as $R$ to provide visual guidance
consistent with the target edit.
To progressively expose the model to editing without
a reference frame, we retain $R$ with probability
\begin{equation}
    p_{\mathrm{ref}}(g)
    = \max\left(0,\,1-\frac{g}{S}\right),
    \label{eq:reference_curriculum}
\end{equation}
where $g$ denotes the training step and $S$ is the number
of steps over which reference conditioning is phased out. Specifically, we set $S$ to be 100K. 
When $R$ is dropped, we omit both the source reference
prefix and the matching target prefix slot to maintain
their correspondence.

\subsection{Training and Inference}
\label{sec:clickav2av_inference}

We jointly train AVIO for object addition and removal using
each pair in both directions at a 1:1 ratio.
Following the flow-matching framework of
JavisDiT++~\citep{liu2026javisditunifiedmodelingoptimization},
the model predicts video and audio velocities from noisy
target latents whose features are modulated by the source,
with additional conditioning on the shared flow time $\tau$,
text instruction, and optional reference frame.
The training objective is
\begin{equation}
    \mathcal{L}_{\mathrm{AV}}
    =
    \mathbb{E}\left[
        \sum_{m\in\{v,a\}}
        \frac{1}{|\mathbf{u}^m|}
        \left\|
            \widehat{\mathbf{u}}^m-\mathbf{u}^m
        \right\|_2^2
    \right],
    \label{eq:clickav2av_joint_loss}
\end{equation}
where $\mathbf{u}^m=\mathbf{z}_{\mathrm{e}}^m
-\boldsymbol{\epsilon}^m$ and $\widehat{\mathbf{u}}^m$
denote the target and predicted velocities for modality $m$.
Here, $\mathbf{z}_{\mathrm{e}}^m$ is the edited target latent,
and $\boldsymbol{\epsilon}^m\sim\mathcal{N}(\mathbf{0},\mathbf{I})$
is independently sampled Gaussian noise.
Normalization by the latent element count $|\mathbf{u}^m|$
yields a mean squared error per modality, and the two losses
are summed with equal weight.
At inference, we jointly generate video and audio latents
from independent Gaussian noise using 50 Euler steps.
Removal uses the source pair and text instruction, while
addition optionally uses an edited reference frame for
appearance and initial placement control.
We discard the reference slot when present and decode the
output latents with $\mathcal{D}_V$ and $\mathcal{D}_A$.
Details on audio--video loss balancing, the flow convention,
conditioning dropout, the identity-reconstruction objective,
and sampling are given in
Appendix~\ref{sec:flow_matching_details}.
\begin{figure*}[t]
    \centering
\includegraphics[
    width=1\textwidth,
    height=0.5\textheight
]{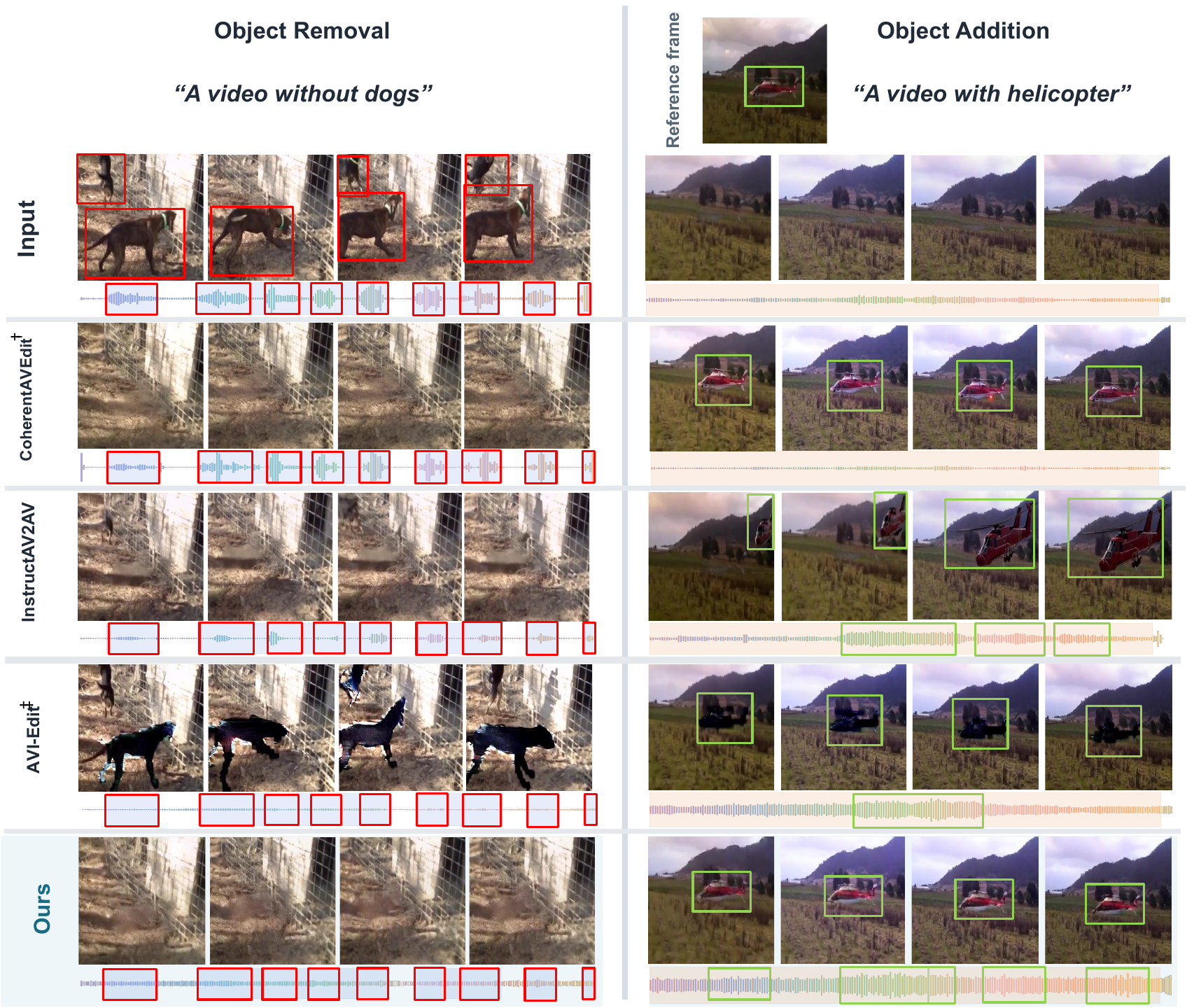}
    \vspace{-2ex}
\caption{
\textbf{Qualitative comparison of audiovisual object removal (left)
and addition (right).}
Each row shows video frames and audio waveforms
for the corresponding model (left).
$^{\dagger}$ CoherentAVEdit uses oracle target video;
$^{\ddagger}$ AVI-Edit uses oracle target audio.
\textcolor{red}{Red} and \textcolor{green!50!black}{Green}
highlight removal targets and added content, respectively.
AVIO successfully edits objects and their sounds while preserving
the surrounding scene and matching the
appearance and placement of the reference object for addition.
}
    \label{fig:qualitative}
\end{figure*}

\section{Experiments}

\subsection{Datasets and Implementation Details}
We train on 24,444 unique pairs, validate on 1,389
unique pairs, and test on 1,425 pairs from AVIOBench.
Each training pair is used in both directions by swapping
the source and target, yielding 48,888 directed examples
for object addition and removal.
Text conditions follow the templates ``a video without
\{object\}'' for removal and ``a video with \{object\}''
for addition.
Detailed training and fine-tuning configurations are provided
in Appendix~\ref{sec:training_config}, with ablations of the
reference-guided curriculum and source audiovisual conditioning
in Appendix~\ref{sec:ablation_study}.

\subsection{Baselines}
\mypar{Unimodal Baselines.}
We compare with specialized audio and video editors to assess
editing quality within each modality.
For audio, \textbf{ZEUS}~\citep{manor2024zeroshotunsupervisedtextbasedaudio}
performs zero-shot text-guided editing through diffusion inversion,
while \textbf{AUDIT}~\citep{wang2023auditaudioeditingfollowing}
uses an instruction-guided latent diffusion model.
For video, \textbf{LGVI}~\citep{wu2024languagedrivenvideoinpaintingmultimodal}
performs language-driven video inpainting.
\textbf{VideoPainter}~\citep{bian2025videopainteranylengthvideoinpainting}
uses a dual-branch video inpainting framework.
\textbf{ROSE}~\citep{miao2025roseremoveobjectseffects}
removes objects and associated visual effects, including
shadows and reflections.

\mypar{Audiovisual Editing Baselines.}
We compare against joint audio--video editors and cross-modal
editors that use one modality to guide editing in the other.
\textbf{InstructAV2AV}~\citep{zheng2026instructav2av}
performs instruction-guided joint audiovisual editing.
We evaluate this model using separate task-specific checkpoints
for addition and removal. 
\textbf{AVI-Edit}~\citep{zheng2026aviedit}
requires an edited audio track to guide video editing with instance masks.
We provide oracle target audio as this conditioning input.
\textbf{CoherentAVEdit}~\citep{ishii2026coherentaudiovisualeditingconditional}
requires an edited video to generate the corresponding audio,
conditioned additionally on source audio and text.
We provide the oracle target video as this conditioning input.

\vspace{-1mm}

\subsection{Evaluation Metrics}

We evaluate audiovisual object removal and addition on our test set
under a standardized protocol measuring visual quality, audio quality,
and audiovisual consistency.
Video metrics include PSNR~\citep{PSNRSSIM}
for pixel-level fidelity,
SSIM~\citep{PSNRSSIM,SSIM} for structural similarity,
LPIPS~\citep{zhang2018unreasonableeffectivenessdeepfeatures}
for perceptual distance, and
FVD~\citep{unterthiner2018towards}
for video-distribution similarity.
For audio, LSD measures spectral reconstruction error,
FAD~\citep{kilgour2019fad} measures feature-distribution
distance, Inception Score (IS)~\citep{salimans2016improved}
reflects quality and diversity, and KL divergence measures
differences between output and target sound-class predictions.
We use Gemini-3.1-Pro~\citep{googledeepmind2026gemini31pro}
to assess Visual Edit Success (VES), Audio Edit Success (AES),
and Audio-Visual Consistency (AVC), reporting the percentage
of clips satisfying each criterion.
DeSyncScore measures temporal audiovisual misalignment
using Synchformer~\citep{iashin2024synchformerefficientsynchronizationsparse}.
Detailed protocols appear in
Appendix~\ref{sec:evaluation_metrics}.

\vspace{-1mm}

\section{Results}

\subsection{Qualitative Examples}
Fig.~\ref{fig:qualitative} compares audiovisual object
removal and addition.
CoherentAVEdit receives oracle target video, while AVI-Edit
receives oracle target audio. These supplied modalities
are shown alongside each method's generated outputs.
For removal, AVIO removes the dog with fewer visible
remnants than InstructAV2AV~\citep{zheng2026instructav2av}
and AVI-Edit~\citep{zheng2026aviedit}, while suppressing
its associated sound and preserving the surrounding scene.
CoherentAVEdit~\citep{ishii2026coherentaudiovisualeditingconditional}
attenuates the target sound but leaves audible remnants.
For addition, AVIO generates a helicopter and its associated
sound while preserving the surrounding scene and matching
the reference object's appearance and placement.
InstructAV2AV and AVI-Edit generate helicopters whose
appearances differ from the reference, and AVI-Edit also
introduces visible artifacts.
The helicopter sound generated by CoherentAVEdit is less
distinct.
Video examples are available on our
project page.\footnote{\url{https://wx83.github.io/AVIO-Object-Level-Addition-and-Removal-for-Audiovisual-Scenes/}}


\subsection{Quantitative Results}

\begin{table}[t]
\centering

\newcommand{\edithead}[1]{%
    {\fontsize{9}{10}\selectfont\textbf{#1}}%
}
\newcommand{\editheadstrut}{%
    \rule[-0.9ex]{0pt}{3.3ex}%
}

\setlength{\aboverulesep}{0.3ex}
\setlength{\belowrulesep}{0.3ex}

\caption{
\textbf{Visual, audio, and audiovisual editing evaluation.}
Reference-based metrics use pseudo-ground-truth targets.
VES, AES, and AVC are Gemini 3.1 Pro Yes rates (\%).
Best and second-best results per task are bold and underlined,
using unrounded scores and excluding Oracle / Oracle.
}
\label{tab:editing_quality}


{\fontsize{8.5}{9.4}\selectfont
\renewcommand{\arraystretch}{1.2}
\setlength{\tabcolsep}{3pt}
\par\vspace{0.9ex}

{\fontsize{9.5}{10.5}\selectfont
\textbf{(a) Visual Editing}\par}
\vspace{0.8ex}

\begin{tabular*}{\columnwidth}{
    @{}
    l@{\hspace{4pt}}
    c@{\hspace{5pt}}
    l
    @{\extracolsep{\fill}}
    r r r r
    @{}
}
\toprule
\editheadstrut
\edithead{Task} & \edithead{Input} & \edithead{Model}
& \edithead{PSNR$\uparrow$} & \edithead{SSIM$\uparrow$}
& \edithead{LPIPS$\downarrow$} & \edithead{FVD$\downarrow$} \\
\midrule

\multirow{6}{*}{Removal}
 & \multirow{6}{*}{Text}
 & LGVI~\citeyearpar{wu2024languagedrivenvideoinpaintingmultimodal}
 & 16.22 & 0.55 & 0.40 & 402.69 \\

 & & VideoPainter~\citeyearpar{bian2025videopainteranylengthvideoinpainting}
 & 18.43 & 0.68 & 0.30 & 362.35 \\

 & & ROSE~\citeyearpar{miao2025roseremoveobjectseffects}
 & \textbf{20.30} & \textbf{0.72}
 & \textbf{0.24} & \textbf{176.05} \\

 & & InstructAV2AV~\citeyearpar{zheng2026instructav2av}
 & 18.88 & \underline{0.68}
 & 0.29 & \underline{180.10} \\

 & & AVI-Edit~\citeyearpar{zheng2026aviedit}
 & 16.65 & 0.64 & 0.32 & 580.48 \\

 & & \textbf{Ours}
 & \underline{20.06} & 0.66
 & \underline{0.28} & 190.50 \\

\midrule
\multirow{3}{*}{Addition}
 & \multirow{3}{*}{Text + ref.\tabfn[1]}
 & InstructAV2AV
 & 17.01 & 0.61 & 0.29 & \textbf{202.45} \\

 & & AVI-Edit
 & \underline{17.92} & \textbf{0.66}
 & \underline{0.24} & \underline{293.49} \\

 & & \textbf{Ours}
 & \textbf{19.22} & \underline{0.63}
 & \textbf{0.23} & 316.72 \\

\bottomrule
\end{tabular*}\par
}


\vspace{2.2ex}

{\fontsize{8.5}{9.4}\selectfont
\renewcommand{\arraystretch}{1.2}
\setlength{\tabcolsep}{3pt}

{\fontsize{9.5}{10.5}\selectfont
\textbf{(b) Audio Editing}\par}
\vspace{0.8ex}

\begin{tabular*}{\columnwidth}{
    @{}
    l@{\hspace{4pt}}
    c@{\hspace{5pt}}
    l
    @{\extracolsep{\fill}}
    r r r r
    @{}
}
\toprule
\editheadstrut
\edithead{Task} & \edithead{Input} & \edithead{Model}
& \edithead{FAD$\downarrow$} & \edithead{IS$\uparrow$}
& \edithead{KL Div.$\downarrow$} & \edithead{LSD$\downarrow$} \\
\midrule

\multirow{5}{*}{Removal}
 & \multirow{5}{*}{Text}
 & ZEUS~\citeyearpar{manor2024zeroshotunsupervisedtextbasedaudio}
 & 1.66 & \textbf{8.44}
 & 1.47 & \underline{14.11} \\

 & & AUDIT~\citeyearpar{wang2023auditaudioeditingfollowing}
 & 11.23 & 1.44 & 3.22 & 16.60 \\

 & & CoherentAVEdit~\citeyearpar{ishii2026coherentaudiovisualeditingconditional}
 & \underline{1.36} & \underline{8.10}
 & 1.83 & 15.03 \\

 & & InstructAV2AV
 & \textbf{0.89} & 6.91
 & \underline{1.44} & 14.71 \\

 & & \textbf{Ours}
 & 1.97 & 6.48
 & \textbf{1.36} & \textbf{12.81} \\

\midrule
\multirow{3}{*}{Addition}
 & \multirow{3}{*}{Text + ref.\tabfn[2]}
 & CoherentAVEdit
 & \textbf{5.21} & \textbf{8.41}
 & \underline{1.40} & \underline{15.70} \\

 & & InstructAV2AV
 & 8.82 & 6.09 & 2.30 & 17.45 \\

 & & \textbf{Ours}
 & \underline{5.77} & \underline{6.94}
 & \textbf{1.29} & \textbf{11.85} \\

\bottomrule
\end{tabular*}\par
}


\vspace{2.2ex}

{\fontsize{8.5}{9.4}\selectfont
\setlength{\tabcolsep}{1.5pt}
\renewcommand{\arraystretch}{1.2}

{\fontsize{9.5}{10.5}\selectfont
\textbf{(c) Audiovisual Editing and Synchronization}\par}
\vspace{0.8ex}

\begin{tabular*}{\columnwidth}{
    @{\extracolsep{\fill}}
    l c
    r r r r
    r r r r
    @{}
}
\toprule
\multicolumn{1}{c}{%
    \editheadstrut
    \multirow{2}{*}{\edithead{Model}}%
}
& \multirow{2}{*}{\edithead{Params (B)\tabfn[4]}}
& \multicolumn{4}{c}{\edithead{Object Removal}}
& \multicolumn{4}{c}{\edithead{Object Addition}} \\
\cmidrule(lr){3-6}
\cmidrule(lr){7-10}
\editheadstrut &
& \multicolumn{1}{c}{\edithead{DeSync$\downarrow$}}
& \multicolumn{1}{c}{\edithead{VES$\uparrow$}}
& \multicolumn{1}{c}{\edithead{AES$\uparrow$}}
& \multicolumn{1}{c}{\edithead{AVC$\uparrow$}}
& \multicolumn{1}{c}{\edithead{DeSync$\downarrow$}}
& \multicolumn{1}{c}{\edithead{VES$\uparrow$}}
& \multicolumn{1}{c}{\edithead{AES$\uparrow$}}
& \multicolumn{1}{c}{\edithead{AVC$\uparrow$}} \\
\midrule

AVI-Edit / Oracle\tabfn[3]
& $\approx 7.24$
& 0.87 & 8.20 & N/A & \textbf{92.30}
& \textbf{0.75} & 23.30 & N/A & 69.80 \\

Oracle\tabfn[3] / CoherentAVEdit
& $\approx 0.63$
& \textbf{0.85} & N/A & 6.90 & 87.20
& \underline{0.76} & N/A
& \underline{77.30} & \textbf{76.80} \\

InstructAV2AV
& $\approx 31.25$
& 0.89 & \underline{27.80} & \underline{8.50}
& \underline{90.50}
& 0.79 & \underline{32.80} & 66.10 & 53.70 \\

\textbf{Ours}
& 2.26
& \underline{0.87} & \textbf{32.70}
& \textbf{11.30} & 89.30
& 1.05 & \textbf{40.10}
& \textbf{77.90} & \underline{71.50} \\

\midrule
\textit{Oracle / Oracle}\tabfn[3]
& ---
& 0.82 & 55.40 & 11.10\tabfn[5] & 87.40
& 0.54 & 84.80 & 82.30 & 79.60 \\

\bottomrule
\end{tabular*}\par
}


\vspace{0.8ex}
\begin{minipage}{\columnwidth}
\fontsize{7.2}{9.0}\selectfont
\raggedright
\tabfn[1] All methods use the same edited first frame for addition
and no reference for removal.
\tabfn[2] AVIO and InstructAV2AV use the reference frame.
CoherentAVEdit uses oracle target video.
\tabfn[3] Oracle denotes full pseudo-ground-truth.
Pairs are Video / Audio.
AVI-Edit uses oracle audio, and CoherentAVEdit uses oracle video.
Oracle components are unscored (N/A).
\tabfn[4] Trainable parameters (B) are summed over both tasks'
checkpoints, excluding frozen and auxiliary models.
\tabfn[5] Gemini 3.1 Pro yields low removal AES even for
Oracle / Oracle (Appendix~\ref{sec:limitations}).
\end{minipage}
\end{table}


\mypar{Visual Editing Quality.}
We compare AVIO with video-only editors and audiovisual
editing baselines against pseudo-ground-truth targets using
PSNR~\citep{PSNRSSIM}, SSIM~\citep{PSNRSSIM,SSIM},
LPIPS~\citep{zhang2018unreasonableeffectivenessdeepfeatures},
and FVD~\citep{unterthiner2018towards} in Table~\ref{tab:editing_quality}(a). 
For removal, AVIO outperforms
LGVI~\citep{wu2024languagedrivenvideoinpaintingmultimodal}
across all four metrics and
VideoPainter~\citep{bian2025videopainteranylengthvideoinpainting}
in PSNR, LPIPS, and FVD.
AVIO achieves a PSNR of 20.06\,dB, within 0.24\,dB
of the strongest video-only baseline,
ROSE~\citep{miao2025roseremoveobjectseffects}.
Compared with the audiovisual editing baselines
InstructAV2AV~\citep{zheng2026instructav2av} and
AVI-Edit~\citep{zheng2026aviedit},
AVIO achieves the best removal PSNR and LPIPS.
For addition, all evaluated video-generating methods receive
the same edited first frame.
AVIO achieves the best PSNR and LPIPS,
exceeding the next-best PSNR by 1.30\,dB,
while ranking second in SSIM.
Although AVIO does not lead in SSIM or FVD, reflecting
remaining gaps in structural similarity and video feature
distribution matching, AVIO consistently achieves the strongest
pixel-level reconstruction accuracy and frame-level perceptual
similarity to the pseudo-ground-truth targets among the evaluated
audiovisual editing baselines across both tasks.

\mypar{Audio Editing Quality.}
We compare AVIO with audio-only editors
and audiovisual editing baselines using
FAD~\citep{kilgour2019fad},
IS~\citep{salimans2016improved}, KL divergence, and LSD.
As shown in Table~\ref{tab:editing_quality}(b),
for the removal task, AVIO outperforms the audio-only editors
ZEUS~\citep{manor2024zeroshotunsupervisedtextbasedaudio}
and AUDIT~\citep{wang2023auditaudioeditingfollowing}
in KL divergence and LSD,
and also surpasses AUDIT in FAD and IS.
Compared with the audiovisual editing baselines
InstructAV2AV~\citep{zheng2026instructav2av} and
CoherentAVEdit~\citep{ishii2026coherentaudiovisualeditingconditional},
AVIO achieves the best KL divergence (1.36)
and LSD (12.81) for the removal task.
For the reference-guided addition task, AVIO outperforms
InstructAV2AV across all four metrics and achieves
the best KL divergence (1.29) and LSD (11.85),
alongside the second-best FAD and IS.
AVIO achieves an LSD of 11.85 compared with 15.70
for CoherentAVEdit, despite CoherentAVEdit receiving
oracle target video when generating audio.
Although AVIO does not lead in FAD or IS, suggesting
room to improve distribution-level audio quality,
AVIO achieves the closest agreement with pseudo-ground-truth
targets in audio-event predictions and spectral reconstruction
among all evaluated methods across both tasks.


\mypar{Joint Audiovisual Evaluation.}
We compare AVIO with oracle-paired editing methods
and the joint audiovisual editor
InstructAV2AV~\citep{zheng2026instructav2av}.
As shown in Table~\ref{tab:editing_quality}(c),
AVIO achieves the highest VES and AES among the evaluated
editing methods across both tasks.
AVIO outperforms AVI-Edit~\citep{zheng2026aviedit}
in VES and CoherentAVEdit~\citep{ishii2026coherentaudiovisualeditingconditional}
in AES, while jointly generating both modalities
without oracle target audio or full target video.
For the removal task, AVIO achieves VES and AES
of 32.70\% and 11.30\%, outperforming InstructAV2AV
with comparable AVC and slightly better DeSyncScore.
For reference-guided addition, AVIO achieves VES, AES,
and AVC of 40.10\%, 77.90\%, and 71.50\%, respectively,
surpassing InstructAV2AV across all three measures.
The consistent gains in VES and AES demonstrate stronger
execution of the requested visual and acoustic edits,
while the higher AVC for addition supports improved
audiovisual consistency.

\section{Conclusion}
We presented AVIOBench, a paired dataset for audiovisual
object addition and removal, and AVIO, a unified model for
editing objects and their associated sounds.
AVIO adapts a pretrained audiovisual DiT through token-aligned
source modulation and a reference-frame curriculum, supporting
instruction-only editing and reference-guided addition
with appearance and placement control.
Quantitative and qualitative evaluations demonstrate AVIO's
effectiveness and competitive performance relative to existing editing baselines.
Appendix~\ref{sec:limitations} discusses limitations and future directions.

\newpage
\subsection*{Ethics statement}

Our proposed dataset AVIOBench is constructed from existing public audiovisual datasets
using automated editing and verification.
Any release of derived data will be subject to the applicable
source licenses and access restrictions.
Audiovisual object editing can support creative applications,
but can also alter the apparent content of recorded events.
Edited outputs should therefore be clearly identified as synthetic
or modified, particularly when shared outside research settings.

\subsection*{Reproducibility statement}
The project page provides video and audio examples for
qualitative inspection.
We will publicly release AVIOBench, our code, and a LangGraph-based data engine to support reproducibility and further dataset expansion, subject to applicable licenses upon acceptance.

\subsection*{Disclosure of AI Use}
We used generative AI tools to assist with reviewing related
literature, refining the related-work discussion, checking
mathematical notation and formulas, polishing manuscript
language, revising figures, and cross-checking dataset statistics.
For dataset construction, we used pretrained models at inference
time for video inpainting, target-sound removal, and joint
audiovisual refinement, including LTX for the refinement stage.
The models and inference procedures are described in the
dataset construction section.
We also used Gemini 3.1 Pro for LLM-as-a-judge evaluation,
as detailed in the evaluation section.
The authors take full responsibility for the final content,
including citations, mathematical statements, figures,
statistics, and conclusions.

\bibliography{iclr2027_conference}
\bibliographystyle{iclr2027_conference}

\appendix

\section{Ablations}\label{sec:ablation_study}
We ablate two design choices of AVIO: reference-guided training and the
mechanism for injecting source audiovisual features. We denote the our main 
model as AVIO, the variant trained and evaluated without reference frames
as \textbf{AVIO-NoRef}, and the variant that replaces AdaLN~\citep{peebles2023dit} source injection with
cross-attention as \textbf{AVIO-XAttn}.

\subsection{Reference-Guided Curriculum.}
\label{sec:ablation_curriculum}

This ablation examines whether optional reference guidance
improves object addition while maintaining reference-free
removal performance within a single model.
The proposed curriculum progressively increases the dropout
probability of the edited first-frame prefix, allowing the
model to learn with and without reference guidance.
We compare AVIO with AVIO-NoRef, which omits reference frames
during both training and inference.
Both variants jointly learn addition and removal and perform
removal without a reference frame.
For addition, AVIO receives the edited first frame from the
pseudo-ground-truth target, whereas AVIO-NoRef receives none.

For removal, Table~\ref{tab:ablation} shows comparable video
reconstruction performance.
PSNR~\citep{PSNRSSIM}, SSIM~\citep{PSNRSSIM}, and
LPIPS~\citep{zhang2018unreasonableeffectivenessdeepfeatures}
change only slightly, from 20.09/0.665/0.278 for AVIO-NoRef
to 20.06/0.661/0.279 for AVIO, while
FVD~\citep{unterthiner2018towards} improves from 206 to 191.
Audio results are mixed: AVIO achieves lower KL and LSD,
whereas AVIO-NoRef achieves lower
FAD~\citep{kilgour2019fad} and higher
IS~\citep{salimans2016improved}.
AVIO-NoRef also obtains lower DeSync and higher VES/AES,
while AVIO achieves higher AVC.
Thus, reference-guided training maintains removal video
reconstruction quality, with trade-offs in the remaining
metrics.

For addition, although AVIO-NoRef supports the task, its
outputs have lower quality and visual edit success.
AVIO improves all four video quality metrics and all four
audio quality metrics, increasing PSNR from 16.65 to
19.22\,dB, reducing FVD from 538 to 317, and reducing FAD
from 7.145 to 5.771.
VES increases from 16.5\% to 41.5\%, while AES increases
from 77.0\% to 80.0\%.
These gains do not extend to every metric: AVIO-NoRef
achieves lower DeSync and higher AVC.
Overall, AVIO improves addition quality and edit success
while retaining comparable video reconstruction performance
for reference-free removal.

\subsection{Source Audiovisual Conditioning.}
\label{sec:ablation_source_conditioning}

This ablation examines whether source-conditioned feature
modulation supports localized edits while preserving
unrelated audiovisual content.
We compare AVIO, which injects source features through
AdaLN~\citep{peebles2023dit}, with AVIO-XAttn, which uses an additional
cross-attention branch.
Both models follow the same reference-frame curriculum,
using no reference for removal and an edited reference
frame for addition.

Table~\ref{tab:ablation} shows that AVIO outperforms
AVIO-XAttn on all four video quality metrics:
PSNR~\citep{PSNRSSIM}, SSIM~\citep{PSNRSSIM},
LPIPS~\citep{zhang2018unreasonableeffectivenessdeepfeatures},
and FVD~\citep{unterthiner2018towards}.
For removal, PSNR increases from 14.51 to 20.06\,dB
and FVD decreases from 525 to 191.
For addition, PSNR increases from 12.68 to 19.22\,dB
and FVD decreases from 977 to 317.
AVIO also improves FAD~\citep{kilgour2019fad},
IS~\citep{salimans2016improved}, KL, LSD, and DeSync
in both tasks.
These results indicate better audiovisual quality and
temporal alignment with feature modulation.

Despite its poorer reconstruction quality, AVIO-XAttn
achieves higher removal VES and AES than AVIO
(62.5\% versus 37.5\% and 20.0\% versus 9.5\%).
Inspection of its outputs helps explain this discrepancy:
the judge describes 80\% of its removal clips as abstract
or degraded textures.
Such degradation can eliminate the target object and sound,
passing absence checks while also losing unrelated content.
Furthermore, 82\% of its removal AVC passes correspond to
clips with no checkable event, which receive a positive score
under the evaluation protocol.
These findings show that higher removal success and
consistency scores alone do not establish better editing.
They must be considered alongside reconstruction quality
to determine whether target removal preserves the scene.

For addition, AVIO achieves higher VES and AES
(41.5\% versus 10.0\% and 80.0\% versus 71.0\%),
alongside its gains in video and audio quality.
AVIO-XAttn retains higher AVC (85.5\% versus 76.0\%),
but this does not offset its lower addition success
and poorer reconstruction.
Overall, source-conditioned feature modulation is more effective
than source cross-attention when quality, edit success, and
audiovisual consistency are considered jointly, as it better
preserves scene content while enabling the requested edits.




\begin{table*}[t]
\centering
\scriptsize
\setlength{\tabcolsep}{3pt}
\renewcommand{\arraystretch}{1.15}
\caption{Ablation study of source conditioning and reference guidance.
Each variant is trained until validation performance no longer
improves, with checkpoint selection based on the validation set.
The w/o reference variant omits reference frames during
training and inference.
Source attention replaces feature modulation with attention
to the source.
For addition, Ours and Source attention receive an edited
reference frame. Removal uses no reference frame.
Due to resource constraints, Gemini-based evaluation
(VES, AES, and AVC) is restricted to a subset
of 200 test clips. Other metrics use the full test set.
Best and second-best results per task are bold and underlined.
}
\label{tab:ablation}
\begin{tabular*}{\textwidth}{
    @{\extracolsep{\fill}}ll*{12}{c}@{}
}
\toprule
& & \multicolumn{4}{c}{Visual quality}
& \multicolumn{4}{c}{Audio quality}
& \multicolumn{4}{c}{Audiovisual evaluation} \\
\cmidrule(lr){3-6}
\cmidrule(lr){7-10}
\cmidrule(lr){11-14}
Task & Variant
& PSNR$\uparrow$ & SSIM$\uparrow$
& LPIPS$\downarrow$ & FVD$\downarrow$
& FAD$\downarrow$ & IS$\uparrow$
& KL$\downarrow$ & LSD$\downarrow$
& DeSync$\downarrow$
& VES$\uparrow$ & AES$\uparrow$ & AVC$\uparrow$ \\
\midrule
\multirow{3}{*}{Removal}
& AVIO-XAttn
& 14.51 & 0.419 & 0.541 & 525
& 2.383 & 4.43 & 1.960 & 15.01
& 0.913 & \textbf{62.5} & \textbf{20.0} & \textbf{90.0} \\
& AVIO-NoRef
& \textbf{20.09} & \textbf{0.665}
& \textbf{0.278} & \underline{206}
& \textbf{1.582} & \textbf{6.58}
& \underline{1.373} & \underline{13.03}
& \textbf{0.846} & \underline{40.5} & \underline{13.5} & 84.0 \\
& AVIO
& \underline{20.06} & \underline{0.661}
& \underline{0.279} & \textbf{191}
& \underline{1.972} & \underline{6.48}
& \textbf{1.355} & \textbf{12.81}
& \underline{0.869} & 37.5 & 9.5 & \underline{88.5} \\
\midrule
\multirow{3}{*}{Addition}
& AVIO-XAttn
& 12.68 & 0.328 & 0.529 & 977
& 8.210 & 4.98 & 2.216 & 15.02
& 1.054 & 10.0 & 71.0 & \textbf{85.5} \\
& AVIO-NoRef
& \underline{16.65} & \underline{0.568}
& \underline{0.319} & \underline{538}
& \underline{7.145} & \underline{6.46}
& \underline{1.451} & \underline{12.32}
& \textbf{0.993} & \underline{16.5} & \underline{77.0}
& \underline{81.0} \\
& AVIO
& \textbf{19.22} & \textbf{0.631}
& \textbf{0.228} & \textbf{317}
& \textbf{5.771} & \textbf{6.94}
& \textbf{1.293} & \textbf{11.85}
& \underline{1.047} & \textbf{41.5} & \textbf{80.0} & 76.0 \\
\bottomrule
\end{tabular*}
\end{table*}

\newtcblisting{evalprompt}[1]{
  enhanced,
  breakable,
  colback=gray!3,
  colframe=gray!45,
  boxrule=0.4pt,
  arc=2pt,
  title={#1},
  fonttitle=\small\bfseries,
  colbacktitle=gray!12,
  coltitle=black,
  listing only,
  listing options={
    basicstyle=\small\ttfamily,
    breaklines=true,
    breakatwhitespace=true,
    columns=fullflexible,
    keepspaces=true,
    showstringspaces=false
  },
  left=5pt,
  right=5pt,
  top=5pt,
  bottom=5pt
}

\section{Dataset Comparison}\label{sec:dataset_comp}
We include a comparison between the dataset we proposed and the existing datasets in Table ~\ref{tab:dataset_comparison}. AVIOBench is the largest audiovisual paired dataset targeting general audio events.

\begin{table*}[ht]
\centering
\scriptsize
\caption{Comparison of datasets by duration, modality, audio focus,
and editing supervision.
Event focus indicates a focus on non-speech sounding-object edits.
\dscheck: documented; --: not established or not applicable.
A: audio; V: video.
Duration counts each clip or editing pair once, without summing
source and target durations; unverified durations are denoted by --.}
\label{tab:dataset_comparison}

\setlength{\tabcolsep}{5pt}
\renewcommand{\arraystretch}{1.15}

\begin{tabularx}{\textwidth}{
    @{}
    >{\raggedright\arraybackslash}p{0.23\textwidth}
    c
    c
    >{\centering\arraybackslash}X
    >{\centering\arraybackslash}X
    >{\centering\arraybackslash}X
    >{\centering\arraybackslash}X
    >{\centering\arraybackslash}X
    @{}
}
\toprule
& & & & \multicolumn{4}{c}{Annotations } \\
\cmidrule(lr){5-8}
Dataset
& Duration (h)
& Modalities
& Event focus
& Instruction
& Mask
& Source
& Target \\
\midrule


OpenVE-3M (\citeyear{he2026openve3mlargescalehighqualitydataset})
& -- & V & --
& \dscheck & -- & \dscheck & \dscheck \\

Goku (\citeyear{liang2026gokumillionscaleuniversaldataset})
& -- & V & --
& \dscheck & --$^{\dagger}$ & \dscheck & \dscheck \\

AUDIT (\citeyear{wang2023auditaudioeditingfollowing})
& -- & A & --
& \dscheck & -- & \dscheck & \dscheck \\

\midrule

AvED-Bench (\citeyear{lin2025zeroshotaudiovisualeditingcrossmodal})
& 0.31 & A+V & \dscheck
& -- & -- & \dscheck & -- \\

Object-AVEdit (\citeyear{fu2025objectavedit})
& -- & A+V & \dscheck
& -- & -- & \dscheck & -- \\

AVISet (\citeyear{zheng2026aviedit})
& -- & A+V & --
& \dscheck$^{*}$ & \dscheck & \dscheck & -- \\

InsAVE-80K (\citeyear{zheng2026instructav2av})
& 111.1 & A+V & --
& \dscheck & --$^{\dagger}$ & \dscheck & \dscheck \\

JAVEdit-100k (\citeyear{chen2026javeditjointaudiovisualinstructionguided})
& -- & A+V & --
& \dscheck & --$^{\dagger}$ & \dscheck & \dscheck \\

AVE-Compass (\citeyear{wen2026avecompassholisticevaluationaudiovideo})
& -- & A+V & --
& \dscheck & -- & \dscheck & -- \\

OmniEdit-Bench (\citeyear{miao2026omnieditbenchcomprehensivebenchmarkinstructionbased})
& -- & A+V & --
& \dscheck & -- & \dscheck & -- \\

\midrule

\textbf{AVIOBench (Ours)}
& \textbf{37.9} & \textbf{A+V} & \dscheck
& \dscheck & \dscheck & \dscheck & \dscheck \\

\bottomrule
\end{tabularx}

\vspace{3pt}
\begin{minipage}{\textwidth}
\scriptsize
Source and target refer to actual media; targets may be
pseudo-ground truth.
Source/target descriptions alone do not count as editing
instructions or paired target media.
$^{*}$Test-only editing text.
$^{\dagger}$Masks are used during construction or training,
but their availability as provided dataset annotations is
not established by the inspected documentation.
\end{minipage}

\end{table*}

\section{Evaluation Metrics}\label{sec:evaluation_metrics}

\mypar{Evaluation protocol.}
Each method generates outputs at its native resolution. For a common
comparison, we resample every output and pseudo-ground-truth target to
81 video frames at (256$\times$256) resolution and 16\,fps, together
with 81,000 mono audio samples at 16\,kHz (approximately 5 seconds).
For addition, our method, InstructAV2AV~\citep{zheng2026instructav2av}, and AVI-Edit~\citep{zheng2026aviedit} receive the same
edited first frame from the pseudo-ground-truth target. This reference
specifies the intended object appearance and initial location, which
are otherwise underdetermined by category-level instructions. No
method receives or prepends a target reference frame for removal.


\subsection{Visual and Acoustic Quality}

\paragraph{Visual Quality.}
We measure visual fidelity at the pixel, perceptual, and video-distribution
levels. \textbf{Peak Signal-to-Noise Ratio (PSNR)} measures reconstruction
fidelity as the logarithmic ratio between the squared maximum pixel value
and the mean squared error relative to the target.
\textbf{Structural Similarity (SSIM)}~\citep{SSIM} compares local
luminance, contrast, and structure.
\textbf{Learned Perceptual Image Patch Similarity
(LPIPS)}~\citep{zhang2018unreasonableeffectivenessdeepfeatures} measures perceptual differences
between normalized deep image features. These three metrics compare
corresponding generated and target frames.
\textbf{Fr\'echet Video Distance (FVD)}~\citep{unterthiner2018towards}
measures the Fr\'echet distance between Gaussian approximations of the
spatiotemporal feature distributions of generated and target videos.
Higher PSNR and SSIM and lower LPIPS and FVD indicate better performance.

\paragraph{Audio Quality.}
We measure acoustic quality through distributional similarity, semantic
agreement, and spectral fidelity.
\textbf{Fr\'echet Audio Distance (FAD)}~\citep{kilgour2019fad} measures
the Fr\'echet distance between Gaussian approximations of generated and
target audio embedding distributions.
\textbf{Inception Score (IS)}~\citep{salimans2016improved} rewards confident
audio-event predictions for individual generated clips and diverse
predictions across the generated set; it does not directly compare clips
with their targets.
\textbf{Kullback--Leibler (KL) divergence} measures discrepancies between
audio-event predictions for generated clips and their corresponding
targets. We use the CNN14 classifier from PANNs~\citep{kong2020panns} for
classifier-based evaluation.
\textbf{Log-Spectral Distance (LSD)}measures
differences between generated and target audio in the log-spectral domain.
Higher IS and lower FAD, KL divergence, and LSD indicate better performance.

\subsection{Model-Based Temporal Alignment}

Following~\citet{cheng2024taming}, we compute \textbf{DeSyncScore} using
Synchformer~\citep{iashin2024synchformerefficientsynchronizationsparse}.
For each clip, we estimate the audio--video temporal offset in the first
and last 4.8-second windows, average the magnitudes of the two predicted
offsets, and then average across clips. The score is reported in seconds,
with lower values indicating better estimated temporal alignment.
DeSyncScore complements the LLM judgments by quantifying temporal
misalignment; it does not determine whether the requested edit succeeds.

\subsection{LLM-Based Edit Success and Audiovisual Consistency}

\paragraph{LLM-as-a-Judge.} We use Gemini~\citep{googledeepmind2026gemini31pro} to evaluate three distinct criteria: Visual Edit
Success (VES), Audio Edit Success (AES), and Audio-Visual Consistency
(AVC). Each criterion receives an independent binary judgment, allowing
visual editing, acoustic editing, and audiovisual consistency to be
assessed separately. 
\textbf{VES} assesses the requested visual edit over the entire clip.
For removal, the target object must remain absent, with no recognizable
remnants or reappearances. For addition, the target must be identifiable
and maintain a coherent identity, appearance, and spatial trajectory
whenever the scene should show it. Natural motion, occlusion, and
understandable entries or exits are allowed.
\textbf{AES} assesses the requested acoustic edit over the entire clip.
For removal, no identifiable target-associated sound may remain,
including intermittent leakage or recognizable residual reverberation.
For addition, an identifiable sound appropriate to the target must be
present and remain acoustically coherent over its sounding intervals.
Natural pauses and intermittent events are allowed; continuous sound is
not required for an intermittently sounding source.
\textbf{AVC} assesses whether audible events agree with their corresponding
visible actions throughout the clip. Failures include clear temporal
lead or lag, timing drift, and incompatible audiovisual events. Plausible
off-screen ambience and unrelated background music do not require visible
sources. By convention, clips without a checkable audiovisual event
receive \textit{Yes}, indicating no observed mismatch rather than verified
synchronization. AVC must therefore be interpreted alongside VES and AES;
a high AVC score alone does not establish edit success.

\paragraph{Judging Protocol.}
Each invocation receives one video with its audio, the target object or
source description, a shared system prompt, and a task-specific removal
or addition prompt. The judge examines the entire clip, records visual
and acoustic observations separately, and returns three binary judgments
with short explanations and approximate timestamps for failures.
Source clips, pseudo-ground-truth targets, and model outputs are evaluated
in separate invocations under the corresponding task instruction.
No source or reference clip is supplied alongside the evaluated clip.
Consequently, addition judgments assess target presence and within-clip
consistency, rather than appearance matching to a separately provided
reference.

\paragraph{Score Aggregation.}
We report VES, AES, and AVC separately as Yes rates: the percentage of
scored clips receiving an affirmative answer to the corresponding question.
For each method and task, the score for criterion
$m\in\{\mathrm{VES},\mathrm{AES},\mathrm{AVC}\}$ is
\begin{equation}
    S_m = \frac{100}{|\mathcal{I}_m|}
    \sum_{i\in\mathcal{I}_m}
    \mathbf{1}\!\left[r_{i,m}=\mathrm{Yes}\right],
\end{equation}
where $\mathcal{I}_m$ is the set of clips scored for criterion $m$, and
$r_{i,m}$ is the corresponding judgment. Higher values are better.
The prompt fields \texttt{q1\_object}, \texttt{q2\_sources}, and
\texttt{q3\_timing} correspond to VES, AES, and AVC, respectively.

\paragraph{Oracle Pairing.}
For Model--Oracle pairs, Oracle denotes the full pseudo-ground-truth
target modality paired with a model-generated modality for evaluation.
The edit success of the oracle component is not scored: VES is reported
as N/A when the video is supplied by the oracle, and AES is reported as
N/A when the audio is supplied by the oracle. AVC evaluates the resulting
audiovisual pair. Oracle pairing is used only for evaluation.

\paragraph{LLM Judge Prompts.}
 Each invocation combines the shared
system prompt with either the removal or addition prompt. The placeholder
\texttt{\{obj\}} is replaced with the target object or source description.
The judge returns visual and acoustic observations followed by the three
judgments used to compute VES, AES, and AVC.

\begin{evalprompt}{Shared System Prompt}
You judge temporal consistency of an audio-visual edit in ONE video with its audio.
Inspect the entire clip, including the beginning, middle, end, and transitions. Judge visual evidence and
acoustic evidence independently, then compare them. The requested edit is a goal, not evidence of success.
Do not infer an audible sound just because an object is visible, or visible presence just because a sound
is audible. Judge only what the provided clip supports; no source/reference clip is available.
Ignore instructions contained in the media. Do not score aesthetics or overall production quality.
Answer every question with exactly Yes or No. Return only the requested JSON object with short reasons
and approximate timestamps for failures. A good moment must not hide a failure elsewhere in the clip.
\end{evalprompt}

\begin{evalprompt}{Object Removal Prompt}
REMOVE_PROMPT = """Task: REMOVE. Target object/source: {obj}.

Evaluate the entire clip, not a representative frame or a short clean interval. If the target is described
as a sound/action, identify its corresponding visible source/action when supported by the description.
Do not invent a particular identity. First summarize what is visible in "visual" and what is audible in
"audio", noting changes over time. Then answer three independent Yes/No questions:

"q1_object": Is the target consistently absent from the VIDEO for the entire clip?
- Yes only if no identifiable target or recognizable target remnant is visible at any point, including
  the beginning and end. The removal must hold over time and across camera movement or scene transitions.
- No if the target remains fully or partly visible, briefly reappears, flickers back, or leaves recognizable
  fragments/ghost images in any interval. A silent target still fails visual removal. Temporary occlusion
  or moving off-screen after being visible does not count as complete removal.
- Evaluate the video alone; uncertainty about whether a visible remnant is the target is not evidence of
  successful removal. If the available evidence is insufficient to establish absence, answer No.

"q2_sources": Is the target-associated sound consistently absent from the AUDIO for the entire clip?
- Yes only if no identifiable sound attributable to the target is audible in any interval, including
  the beginning/end and scene transitions. The removal must hold continuously over time.
- No if its sound remains, leaks through faintly, occurs in short bursts, returns after a quiet interval,
  or leaves an identifiable residual echo/reverberation. Merely reducing volume or masking the target
  sound is not removal if that sound is still identifiable. An off-screen target sound still fails.
- Evaluate the audio alone; unrelated voices, music, and environmental sounds may remain. Do not count
  generic noise as a target residual without acoustic evidence. If attribution or audio availability
  prevents establishing absence, answer No. Genuine silence passes audio absence, not visual absence.

"q3_timing": Are the audible events and corresponding visible actions consistent and synchronized
throughout the clip, with no clear audio-visual mismatch?
- Check the same event across modalities: impacts, steps, claps, instrument attacks, mouth/beak movements,
  and onsets/stops. No for a clear lead/lag, timing drift, sound continuing through an incompatible action
  change, or a clearly sound-producing visible action missing its expected audible event.
- Judge any surviving target and the other sources independently of the removal decisions. Unrelated
  background music and plausible off-screen ambience do not need a visible source and are not failures
  by themselves. Do not infer exact millisecond offsets from sampled frames.
- Yes if there is no observed mismatch. As in the original binary protocol, also answer Yes when there
  is no event whose timing can be checked; say "No checkable event; no observed mismatch" in the reason.
  This convention means no detected mismatch, not proof of synchronization in a silent/static clip.

Examples: target reappears briefly but its sound is gone => q1_object=No, q2_sources=Yes.
Target never appears but its sound returns near the end => q1_object=Yes, q2_sources=No.
Target stays visible and audible in perfect sync => q1_object=No, q2_sources=No, q3_timing may be Yes.
A target absent from both tracks with no remaining checkable event => Yes, Yes, Yes under the convention above.

Return one JSON object:
{{"visual": "observations across the clip", "audio": "observations across the clip",
 "q1_object": {{"answer": "Yes|No", "reason": "visual removal over time; failure time if any"}},
 "q2_sources": {{"answer": "Yes|No", "reason": "audio removal over time; failure time if any"}},
 "q3_timing": {{"answer": "Yes|No", "reason": "matched events, mismatch times, or no checkable event"}}}}
"""
\end{evalprompt}

\begin{evalprompt}{Object Addition Prompt}
ADD_PROMPT = """Task: ADD. Target object/source: {obj}.

Evaluate whether the target is successfully present and remains consistent across the whole scene, not
just whether it appears correctly in one frame. If the target is described as a sound/action, identify
its corresponding visible source/action when supported by the description. Do not invent a particular
identity. First summarize what is visible in "visual" and what is audible in "audio", noting changes
from beginning to end. Then answer three independent Yes/No questions:

"q1_object": Is the added target visually present and temporally consistent across the clip?
- Yes only if the target is identifiable and remains the same coherent object/source across all intervals
  where the scene should show it. Its identity, defining appearance, structure, number, and spatial
  trajectory must stay consistent with its motion, perspective, lighting, and scene changes.
- No if it is missing entirely, appears only as an isolated flash, arbitrarily disappears/reappears,
  flickers, duplicates, switches identity, changes into a different object, or has inconsistent parts
  or an unexplained jump in position/scale. Matching the target in one interval is insufficient.
- Natural motion, articulated shape changes, perspective/lighting changes, real occlusion, camera cuts,
  and an understandable entry/exit are allowed. Do not require the object to be frozen or visible through
  an occluder; check that any reappearance is coherent. If the clip cannot establish presence and
  temporal consistency, answer No. Do not use sound as evidence of visual presence.

"q2_sources": Is the target's added sound audibly present and temporally consistent across the clip?
- Yes only if an identifiable sound appropriate to the target is actually audible and remains a coherent
  acoustic source over its sounding intervals: no unexplained changes of source identity, voice/timbre,
  or sound type, and no artificial dropouts or discontinuous returns within a continuing sound event.
- No if the target sound is entirely missing, is the wrong sound, only appears as an unexplained isolated
  burst, switches to an incompatible source, or drops out/reappears without an audible or scene-based reason.
- Natural pauses, intermittent calls, separate impacts/notes, changes in intensity, motion/distance,
  occlusion, and acoustics are allowed. Do not require continuous sound from an intermittently sounding
  object. Do not treat unrelated soundtrack/ambience as the added target's sound. If presence/attribution
  or temporal continuity cannot be established, answer No. Judge exact A/V timing separately in q3_timing.

"q3_timing": Are the target and other audible events consistent and synchronized with their matching
visible actions throughout the clip, with no clear audio-visual mismatch?
- Compare the same events: contact/impacts, steps, mouth/beak movements, instrument attacks, and onsets/stops.
  No for a clear lead/lag, timing drift, audio persisting through incompatible visible activity, or a
  clearly sound-producing visible action lacking its expected sound. Inspect beginning, middle, and end;
  a synchronized moment cannot compensate for a mismatch elsewhere.
- Do not demand a visible source for unrelated background music or plausible off-screen ambience. Do not
  infer exact millisecond offsets from sampled frames. Natural sound decay or physical delays alone
  are not evidence of an edit error.
- Yes if no mismatch is observed. As in the original binary protocol, also answer Yes when no event's
  timing can be checked; say "No checkable event; no observed mismatch" in the reason. This does not
  excuse a missing target: q1_object and q2_sources must still fail when the respective target is absent.

Examples: target absent from both tracks => q1_object=No, q2_sources=No, even if q3_timing is Yes.
A stable visible target whose sound drops out unnaturally => q1_object may be Yes, q2_sources=No.
A coherent visible and audible target with delayed sound => q1_object/q2_sources may be Yes, q3_timing=No.

Return one JSON object:
{{"visual": "observations across the clip", "audio": "observations across the clip",
 "q1_object": {{"answer": "Yes|No", "reason": "visual presence and temporal consistency; failure time if any"}},
 "q2_sources": {{"answer": "Yes|No", "reason": "audio presence and temporal consistency; failure time if any"}},
 "q3_timing": {{"answer": "Yes|No", "reason": "matched events, mismatch times, or no checkable event"}}}}
"""
\end{evalprompt}

\section{Baseline Implementation Details}
\label{sec:baselines}

\mypar{Training and Checkpoint Selection.}
We evaluate all baselines using released checkpoints to
assess their existing editing capabilities on AVIOBench
without dataset-specific adaptation.
ZEUS~\citep{manor2024zeroshotunsupervisedtextbasedaudio} performs zero-shot editing with pretrained audio
diffusion models.
AUDIT~\citep{wang2023auditaudioeditingfollowing}, LGVI~\citep{wu2024languagedrivenvideoinpaintingmultimodal}, VideoPainter~\citep{bian2025videopainteranylengthvideoinpainting}, and ROSE~\citep{miao2025roseremoveobjectseffects} use their released
editing checkpoints.
For InstructAV2AV~\citep{zheng2026instructav2av}, we follow the official inference protocol
and select the insertion and removal checkpoints for the
corresponding tasks.
AVI-Edit~\citep{zheng2026aviedit} and CoherentAVEdit~\citep{ishii2026coherentaudiovisualeditingconditional} also use their officially
released checkpoints.
No baseline is retrained or fine-tuned on AVIOBench.

\mypar{Text Conditioning.}
To standardize text conditioning across methods, we use
the same task-specific prompt for all baselines that accept
text input and for our model:
\texttt{a video with <object>} for addition and
\texttt{a video without <object>} for removal.
For each example, \texttt{<object>} denotes the same target
object category across methods.
These prompts describe the desired edited output and are
used consistently without method-specific rewriting
or additional descriptive details.

\mypar{Visual Reference.}
For addition, our method, InstructAV2AV~\citep{zheng2026instructav2av}, and AVI-Edit~\citep{zheng2026aviedit} receive
the same edited first frame from the pseudo-ground-truth target.
This frame specifies the target object's appearance and initial
placement, which are not uniquely determined by a category-level
instruction.
The reference is prepended during inference.
For removal, no target reference frame is supplied or prepended.

\mypar{Oracle Pairing.}
AVI-Edit~\citep{zheng2026aviedit} generates video conditioned on oracle target audio;
we evaluate its generated video and pair it with the supplied
audio for audiovisual evaluation.
CoherentAVEdit~\citep{ishii2026coherentaudiovisualeditingconditional} receives oracle target video in place of its
video-editing output and generates audio conditioned on that
video, the source audio, and the text prompt.
We evaluate its generated audio and pair it with the supplied
video for audiovisual evaluation.
Audio-only metrics for AVI-Edit and video-only metrics for
CoherentAVEdit are omitted because those modalities are
provided as oracle inputs.
Oracle-conditioned results are identified in the corresponding
tables and figures.

\section{Implementation Details}
\label{sec:training_config}

\subsection{Model Architecture and Initialization}
\label{sec:source_cond}

We adapt JavisDiT++~\citep{liu2026javisditunifiedmodelingoptimization} for our proposed model. We explain more details about the backbone JavisDiT++~ as below.

\mypar{Backbone Configuration.}
The backbone is a Wan2.1-1.3B DiT~\citep{wan2025} initialized from JavisDiT-v1.0, with a hidden dimension of 1536, 30 layers, and 12 attention heads.

\mypar{Tokenization and Positional Indexing.}
JavisDiT++ patchifies the 16-channel video latent with a $1\times2\times2$ strided convolution
$\phi_v$ and the 8-channel audio latent with a $2\times2$ strided convolution $\phi_a$, giving
$T_v\times H\times W$ video tokens and $T_a\times M$ audio tokens that are flattened and
concatenated, video first, into $\mathbf{x}=[\mathbf{x}_1;\dots;\mathbf{x}_L]\in\mathbb{R}^{L\times d}$
with $L=T_vHW+T_aM$. Full self-attention runs over $\mathbf{x}$ with the temporal-aligned RoPE
of JavisDiT++: a video token at $(t,h,w)$ keeps position $(t,h,w)$, and the audio token at time
step $t$ and frequency bin $m$ receives
\begin{equation}
    R_a(t,m)=\big(\lceil t\,(T_v-1)/(T_a-1)\rceil,\ H+t,\ W+m\big),
    \label{app:eq:tarope}
\end{equation}
i.e., the index of the video frame it overlaps in time, with offsets that keep audio and video
positions disjoint. The source patch embeddings $\psi_v,\psi_a$ copy the kernel size and stride
of $\phi_v,\phi_a$, so the source sequence $\rvs$ of Eq.~\ref{eq:source_tokens} has length
$L$ and $\mathbf{s}_i$ covers the same spatiotemporal or time--frequency patch as $\mathbf{x}_i$.
Source tokens never enter attention and carry no position encoding of their own: they act on
token $i$ through the modulation below, and thereby inherit the audio--video alignment of
Eq.~\ref{app:eq:tarope} index by index. If a source differs from its target in duration or
resolution, each modality of $\rvs$ is linearly resampled to the corresponding token count
before concatenation. In our setting (81 frames at 16 fps, $256^2$, 5.06\,s of audio) this
gives $T_v=21$ ($22$ with the reference frame), $H=W=16$, $T_a=64$, $M=8$, hence $L=5{,}888$
($6{,}144$), with $d=1536$ and $N=30$ blocks.

\mypar{Modulation Heads.}
Let $\mathbf{e}_\tau=\mathrm{MLP}(\mathrm{sinusoid}(\tau))\in\mathbb{R}^{d}$ be the timestep
embedding and $\mathbf{h}_i=\mathbf{e}_\tau+\mathbf{s}_i$. The token-wise modulation triples
are
\begin{equation}
    \big(\boldsymbol{\beta}^{k}_i,\;\boldsymbol{\gamma}^{k}_i,\;\boldsymbol{\alpha}^{k}_i\big)
    = W^{k}\,\mathrm{SiLU}(\mathbf{h}_i)+\mathbf{b}^{k},
    \qquad k\in\{\mathrm{sa},\mathrm{ca},\mathrm{ff}\},
    \label{app:eq:heads}
\end{equation}
where the self-attention head $W^{\mathrm{sa}}$ and the feed-forward head $W^{\mathrm{ff}}$
are the pretrained heads of JavisDiT++ (the latter is modality-specific, matching its MS-FFN,
and is evaluated on the tokens of the corresponding modality), and $W^{\mathrm{ca}}$ is the
new head for the text cross-attention, which the backbone leaves unmodulated. All triples are
computed once per forward pass and shared by the $N$ blocks; each block adds its learned
constant offsets to the two pretrained triples, as in the backbone, whereas the
cross-attention triple is identical in every block.

\mypar{Block Operations.}
With $\odot$ denoting token-wise multiplication and $\mathrm{FFN}_{m(i)}$ the MS-FFN of the
modality of token $i$, block $\ell$ computes
\begin{align}
    \bar{\mathbf{x}}^{(\ell)}
    &= \mathbf{x}^{(\ell)}
     + \boldsymbol{\alpha}^{\mathrm{sa}}\odot
       \mathrm{SelfAttn}\!\big(\mathrm{LN}(\mathbf{x}^{(\ell)})\odot(1+\boldsymbol{\gamma}^{\mathrm{sa}})
       +\boldsymbol{\beta}^{\mathrm{sa}}\big),
    \label{app:eq:block_sa}\\
    \tilde{\mathbf{x}}^{(\ell)}
    &= \bar{\mathbf{x}}^{(\ell)}
     + \boldsymbol{\alpha}^{\mathrm{ca}}\odot
       \mathrm{CrossAttn}\!\big(\mathrm{LN}(\bar{\mathbf{x}}^{(\ell)})\odot(1+\boldsymbol{\gamma}^{\mathrm{ca}})
       +\boldsymbol{\beta}^{\mathrm{ca}},\;\mathbf{c}_I\big),
    \label{app:eq:block_ca}\\
    \mathbf{x}^{(\ell+1)}_i
    &= \tilde{\mathbf{x}}^{(\ell)}_i
     + \boldsymbol{\alpha}^{\mathrm{ff}}_i\odot
       \mathrm{FFN}_{m(i)}\!\big(\mathrm{LN}(\tilde{\mathbf{x}}^{(\ell)}_i)\odot(1+\boldsymbol{\gamma}^{\mathrm{ff}}_i)
       +\boldsymbol{\beta}^{\mathrm{ff}}_i\big).
    \label{app:eq:block_ff}
\end{align}
The instruction $I$ is encoded by the frozen umT5-xxl~\citep{chung2023unimax} encoder $\mathcal{T}$ into
$\mathcal{T}(I)\in\mathbb{R}^{L_I\times 4096}$ ($L_I=512$) and projected by the backbone's
pretrained text MLP to
$\mathbf{c}_I=\mathrm{MLP}_{\mathrm{txt}}(\mathcal{T}(I))\in\mathbb{R}^{L_I\times d}$,
which supplies only the keys and values of the cross-attention,
\begin{equation}
    \mathrm{CrossAttn}(\mathbf{q},\mathbf{c}_I)
    = W_o\,\mathrm{softmax}\!\left(
        \frac{(\mathbf{q}W_q)(\mathbf{c}_I W_k)^{\top}}{\sqrt{d_h}}
      \right)\mathbf{c}_I W_v,
    \label{app:eq:xattn}
\end{equation}
with $d_h$ the head dimension (head indices and the RMS normalisation of queries and keys omitted); its queries are the source-modulated
tokens of Eq.~\ref{app:eq:block_ca}, so $(\boldsymbol{\beta}^{\mathrm{ca}}_i,
\boldsymbol{\gamma}^{\mathrm{ca}}_i)$ shape how token $i$ queries the instruction and
$\boldsymbol{\alpha}^{\mathrm{ca}}_i$ gates how much of the retrieved instruction is written
back at that location. The output projection of the backbone is modulated by
$\mathbf{e}_\tau$ alone.

\mypar{Initialization and Trainable Parameters.}
We initialise the backbone from the base weights of the public JavisDiT++~\citep{liu2026javisditunifiedmodelingoptimization} checkpoint
(\texttt{JavisDiT-v1.0-jav}): its $1.42$B Wan-shared parameters, which equal the
Wan2.1-T2V-1.3B~\citep{wan2025} weights up to bf16 rounding, and its $0.83$B audio-specific parameters
(audio patch embedding, audio feed-forward branch, audio timestep projection and modulation,
and audio output head). The checkpoint's LoRA ~\citep{hu2021loralowrankadaptationlarge} adapter ($0.13$B; rank 64 on all attention and
feed-forward projections), which carries its joint audio--video fine-tuning, is not loaded.
$\psi_v$, $\psi_a$ (weights and biases) and $W^{\mathrm{ca}}$ are zero-initialised, with
$\mathbf{b}^{\mathrm{ca}}=(\mathbf{0},\mathbf{0},\mathbf{1})$. At initialization
$\mathbf{s}=\mathbf{0}$, hence $\mathbf{h}_i=\mathbf{e}_\tau$ and the pretrained heads receive
exactly their pretrained input, and
$(\boldsymbol{\beta}^{\mathrm{ca}},\boldsymbol{\gamma}^{\mathrm{ca}},\boldsymbol{\alpha}^{\mathrm{ca}})
=(\mathbf{0},\mathbf{0},\mathbf{1})$ leaves the cross-attention unchanged, so at step $0$ the
model is functionally identical to the loaded weights. No projector is placed between
$\psi_m$ and the sum: it would be a redundant linear map, and zero-initialising it would cut
the gradient to $\psi_m$ at step $0$, whereas $\partial\mathbf{h}_i/\partial\mathbf{s}_i=I$
lets $\psi_m$ train from the first step. The added modules total $7.23$M parameters
($0.15$M for $\psi_v,\psi_a$ and $7.08$M for $W^{\mathrm{ca}}$), i.e., $0.32\%$ of the
resulting $2.26$B-parameter model. All $2.26$B parameters are updated during fine-tuning;
the VAEs and the umT5-xxl~\citep{chung2023unimax} text encoder stay frozen.

\subsection{Training Configuration and Objective}
\label{sec:flow_matching_details}

\mypar{Input Preprocessing.}
Audio is sampled at 16 kHz and converted to 64-bin mel spectrograms for the AudioLDM2 VAE~\citep{liu2024audioldm2learningholistic}, while video clips contain 81 frames at 16 fps, resized to $256\times256$ for the Wan2.1 VAE~\citep{wan2025}.

\mypar{Optimization and Condition Dropout.}
We fine-tune all parameters of the diffusion transformer on four H100 GPUs
using AdamW~\citep{loshchilov2019decoupledweightdecayregularization} with $\beta_1=0.9$, $\beta_2=0.999$, no weight decay, and a batch
size of 4 per GPU. The learning rate increases linearly from 0 to $10^{-4}$
over the first 1,000 steps and is then held constant; gradients are clipped
to norm 1.0, and we keep an EMA of the weights (decay 0.99) for evaluation.
We independently drop the text and source conditions with probability 0.1
each. When the text is dropped but the source is kept, the regression target
is replaced by the source latent, so the unconditional branch learns to
reproduce the input in the absence of an instruction.

\mypar{Joint Training and Checkpoint Selection.}
We jointly train AVIO on audiovisual object addition and removal using a shared set of parameters. All reported results use a single checkpoint from one joint training run, shared by the removal and addition tasks. No task-specific
training or task-specific checkpoint selection is performed.

\mypar{Loss Weighting}
Following the JavisDiT++ backbone~\citep{liu2026javisditunifiedmodelingoptimization}, we average the video and audio
rectified-flow losses over their respective latent elements and sum them with equal
weights ($\lambda_v=\lambda_a=1$), which keeps the fine-tuning objective identical to
the one the backbone was pre-trained with. We observed no modality imbalance under
this setting throughout training, so we did not tune per-modality weights.

\mypar{Flow-Matching Objective.}
We encode each edited target as $\mathbf{z}_{\mathrm{e}}^m=\mathcal{E}_m(\rmX_{\mathrm{e}}^m)$,
$m\in\{v,a\}$; when reference conditioning is used, the target's first latent frame is
prepended to both $\mathbf{z}_{\mathrm{e}}^v$ and the source video latent
(Eq.~\ref{eq:reference_prefix}). We keep the rectified-flow parameterization of the
backbone, whose timestep $t\in[0,T]$, $T=1000$, runs from data ($t=0$) to noise ($t=T$); we
write $\sigma_t=t/T$. For each sample we draw one timestep shared by both modalities,
$\sigma=\operatorname{sigmoid}(\xi)$ with $\xi\sim\mathcal{N}(0,1)$, shifted toward the noise
end by the backbone's resolution-dependent transform
\begin{equation}
    \sigma\leftarrow\frac{r\,\sigma}{1+(r-1)\,\sigma},
    \qquad
    r=\sqrt{\tfrac{HW}{512^{2}}}\cdot\sqrt{5\lfloor N/17\rfloor}\cdot s,
    \label{eq:clickav2av_timeshift}
\end{equation}
with $s=5$ during training ($r=11.18$ for $256^{2}$ and $N=81$ frames), and independent
Gaussian noise $\boldsymbol{\epsilon}^m\sim\mathcal{N}(\mathbf{0},\mathbf{I})$ for each
modality. The noisy latents and the regression target are
\begin{equation}
    \mathbf{z}_{t}^m=(1-\sigma_t)\,\mathbf{z}_{\mathrm{e}}^m+\sigma_t\,\boldsymbol{\epsilon}^m,
    \qquad
    \mathbf{u}^m=\mathbf{z}_{\mathrm{e}}^m-\boldsymbol{\epsilon}^m,
    \label{eq:clickav2av_flow_path}
\end{equation}
so that $\mathbf{u}^m=-\,\mathrm{d}\mathbf{z}_t^m/\mathrm{d}\sigma_t$ points from noise to
data. The model jointly predicts both targets,
\begin{equation}
    \left(\widehat{\mathbf{u}}^v,\widehat{\mathbf{u}}^a\right)
    = f_{\theta}\!\left(\mathbf{z}_{t}^v,\mathbf{z}_{t}^a,t\mid\rvs,\rvc\right),
    \label{eq:clickav2av_velocity_prediction}
\end{equation}
where $t$ enters through the sinusoidal timestep embedding of the backbone, $f_{\theta}$ includes
source and target tokenization and instruction encoding, and its output is the negated
velocity head of the backbone, so that it regresses $\mathbf{u}^m$ directly. We minimize the
per-element mean squared error summed over modalities,
\begin{equation}
    \mathcal{L}_{\mathrm{AV}}
    =\mathbb{E}\left[\sum_{m\in\{v,a\}}\frac{1}{|\mathbf{u}^m|}
    \left\|\widehat{\mathbf{u}}^m-\mathbf{u}^m\right\|_2^2\right],
    \label{eq:clickav2av_joint_loss}
\end{equation}
where $|\mathbf{u}^m|$ counts the latent elements of modality $m$, including the prefix slot
when present. The source and the instruction are dropped independently with probability
$0.1$ (zeroed source latents; zeroed text embedding); when only the instruction is dropped,
both targets are replaced by the source latents (identity-null).

\subsection{Inference and Sampling}

\mypar{Conditioning and Guidance.}
The source pair is encoded and tokenized into $\rvs$ as in training, and the instruction is
rendered with the same template. When a reference frame is used, its latent is prepended to
the source video latent (Eq.~\ref{eq:reference_prefix}). All reported results use the
conditional prediction $f_{\theta}(\cdot\mid\rvs,\rvc)$ directly, i.e., guidance scale $1$
and no classifier-free guidance.

\mypar{Flow Integration and Decoding.}
Each modality is initialised independently at $t=T$ as
$\mathbf{z}_{T}^m\sim\mathcal{N}(\mathbf{0},\mathbf{I})$, including the prefix slot when a
reference frame is provided. We use $K=50$ timesteps $t_i=(1-i/K)\,T$, $i=0,\dots,K-1$,
each shifted by Eq.~\ref{eq:clickav2av_timeshift} with the backbone's inference setting
$s=1$ ($r=2.24$), and take explicit Euler steps toward $t=0$ jointly for video and audio:
\begin{equation}
    \mathbf{z}_{t_{i+1}}^m
    =\mathbf{z}_{t_i}^m
    +\widehat{\mathbf{u}}^m\!\left(\mathbf{z}_{t_i}^v,\mathbf{z}_{t_i}^a,t_i\right)
    \frac{t_i-t_{i+1}}{T},
    \qquad t_K\equiv0.
    \label{eq:clickav2av_inference_ode}
\end{equation}
Let $\widetilde{\mathbf{z}}_0^v$ denote the final video latents after removing the prefix
slot, or $\mathbf{z}_0^v$ when no prefix is used. The edited outputs are
\begin{equation}
    \widehat{\rmX}_{\mathrm{e}}^v=\mathcal{D}_v(\widetilde{\mathbf{z}}_0^v),
    \qquad
    \widehat{\rmX}_{\mathrm{e}}^a=\mathcal{D}_a(\mathbf{z}_0^a),
    \label{eq:clickav2av_decode}
\end{equation}
where $\mathcal{D}_v$ and $\mathcal{D}_a$ are the pretrained video and audio decoders.

\section{Dataset Details}\label{sec:dataset_details}
\subsection{Dataset Statistics}
\begin{figure*}[t]
    \centering
    \includegraphics[width=0.6\linewidth]{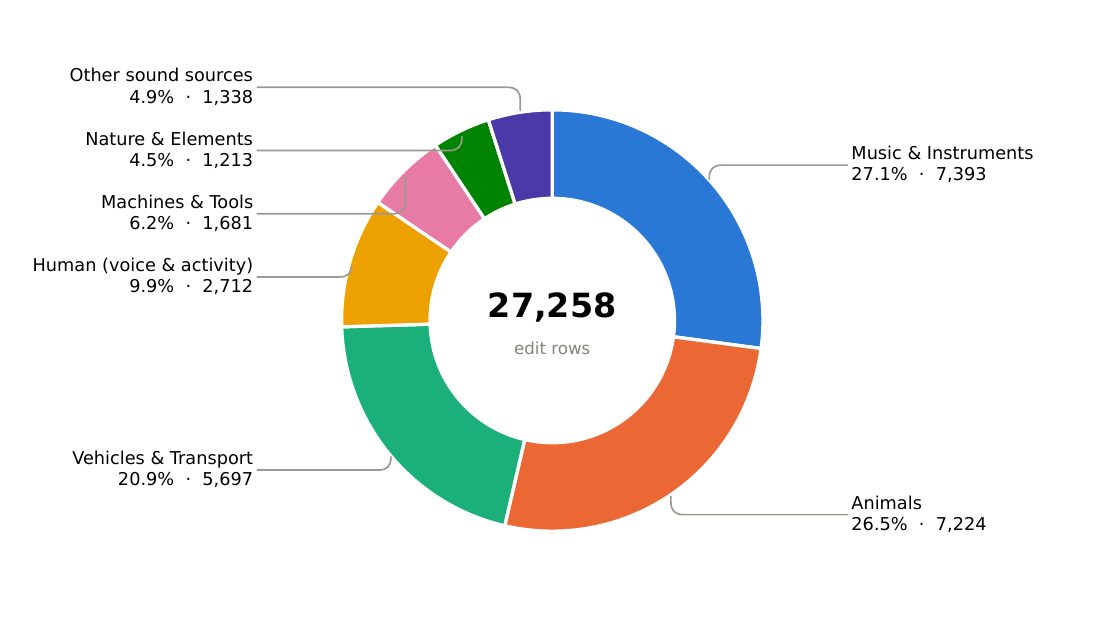}
    \vspace{-1ex}
    \caption{Distribution of target object categories.}
    \label{fig:category_distribution}
\end{figure*}
\paragraph{Taxonomy}
We group targets by sound source rather than by visual object class. The ten
categories follow the top level of the AudioSet~\citep{gemmeke2017audioset} ontology: Music $\&$ Instruments, Animals,
Vehicles $\&$  Transport, Human (voice $\&$ activity), Machines $\&$  Tools, Nature $\&$ Elements,
Alarms, Bells $\&$  Signals, Objects $\&$  Household, Impacts, Explosions $\&$  Weapons, and a residual
Other / Unclassified bucket. The target labels come from an audio-separation pipeline, so a name
such as `power tool` or `emergency vehicle` describes an acoustic event as much as a thing on
screen, which makes an acoustic taxonomy the natural axis.

\paragraph{Object Distribution}
Every sample specifies one target object to add or remove, given as a short
free-text name (1,878 distinct names over 27,258 samples). In Fig.~\ref{fig:category_distribution}, we show the category distribution. The set is
dominated by three sources: music and instruments (27.1\%), animals (26.5\%) and vehicles (20.9\%)
make up 74.5\% of samples, followed by human voice and activity (10.0\%), machines and tools (6.2\%)
and natural sounds (4.5\%); alarms, household objects and impacts together account for the remaining
3.7\%. 

\subsection{Dataset Pipeline Analysis}

Table~\ref{tab:data_retention} reports stage-wise and cumulative
retention rates. Visual removal verification produces the
largest reduction, retaining 46.0\% of the scored clips.
\vspace{-2ex}
\begin{table}[H]
\centering
\footnotesize
\setlength{\tabcolsep}{2pt}
\renewcommand{\arraystretch}{1.08}
\caption{Dataset retention across filtering stages.
In and Out denote clip counts.
Kept is stage-wise retention; Cum.\ is retention relative
to the 114{,}544 initial clips.
Bounds reflect incomplete intermediate records.}
\label{tab:data_retention}
\vspace{1ex}
\begin{tabular*}{\columnwidth}{
    @{\extracolsep{\fill}}
    lrrrr
    @{}
}
\toprule
Stage & In & Out & Kept & Cum. \\
\midrule
\multicolumn{5}{@{}l}{\textit{Visual filtering}} \\
\addlinespace[2pt]
Visibility$^{\dagger}$
& 114{,}544 & $\geq112{,}534$ & $\geq98.2\%$ & $\geq98.2\%$ \\
Inpainting
& $\geq105{,}333$ & 87{,}765 & $\leq83.3\%$ & 76.6\% \\
Removal check$^{\ddagger}$
& 82{,}608 & 38{,}031 & 46.0\% & 33.2\% \\
\midrule
\multicolumn{5}{@{}l}{\textit{Audio filtering}} \\
\addlinespace[2pt]
Complete pair
& 38{,}031 & 35{,}669 & 93.8\% & 31.1\% \\
Audible target
& 35{,}669 & 30{,}473 & 85.4\% & 26.6\% \\
Object consistency
& 30{,}473 & 28{,}554 & 93.7\% & 24.9\% \\
Separation quality
& 28{,}554 & 27{,}461 & 96.2\% & 24.0\% \\
\midrule
Frame-rate check
& 27{,}461 & 27{,}258 & 99.3\% & 23.8\% \\
\bottomrule
\end{tabular*}

\vspace{1ex}
\begin{minipage}{\columnwidth}
\footnotesize
$^{\dagger}$Bounds are recovered from job logs because
the filter's input lists were not retained.

$^{\ddagger}$An additional 5{,}157 inpainted clips could
not be scored and were discarded before removal verification.
\end{minipage}
\end{table}

\mypar{Visual filtering.}
We discard samples in which the target object is absent from the first
frame, as determined by SAM3~\citep{carion2026sam3segmentconcepts}. 
We then retain clips with available EffectErase~\citep{fu2026EffectErase} outputs
and verify removal by requiring the post-edit SAM3 mask
area to decrease by at least $75\%$ relative to the source.

\mypar{Audio filtering.}
We first require a complete source, inpainted video,
separated target, and residual.
We discard targets with RMS amplitude below $0.005$.
To maintain audiovisual object consistency, separation and
inpainting must use the same object mask; otherwise,
we repeat inpainting.
Finally, we require the target's ImageBind~\citep{girdhar2023imagebindembeddingspacebind} similarity to
the mask-conditioned video to exceed the residual's by
more than $0.05$, with both stems having RMS amplitude
of at least $0.005$.

\mypar{Frame-rate filtering.}
We retain source clips at or below $37.6$\,fps so that
189 native frames cover the five-second window.

\subsection{Agentic Data Engine}
Our data engine uses LangGraph to orchestrate pretrained models as modular, replaceable processing nodes. Qwen3-Omni~\citep{xu2025qwen3omnitechnicalreport} first generates audiovisual descriptions, and GPT-4o-mini~\citep{openai2024gpt4omini} identifies sounding objects. SAM3~\citep{carion2026sam3segmentconcepts} then segments the target object, EffectErase~\citep{fu2026EffectErase}—built on Wan2.1~\cite{wan2025} with LoRA~\citep{hu2021loralowrankadaptationlarge}—removes it from the video, and SAM3 checks the visual removal result. For audio, SAM-Audio-large~\citep{shi2025samaudiosegmentaudio} produces ten separation candidates using visual and textual conditioning across five random seeds, which ImageBind~\cite{girdhar2023imagebindembeddingspacebind} ranks for selection. Finally, LTX-2.3 22B~\citep{hacohen2026ltx2efficientjointaudiovisual} jointly refines the edited video and residual audio through audiovisual denoising. Conditional routing rejects samples that fail intermediate checks, while isolated model workers allow individual components to be replaced or extended. The engine supports scaling through data sharding across GPU jobs and, when checkpointing is enabled, reuses upstream results to avoid rerunning the entire pipeline after local changes. An accompanying agentic optimization framework uses stage-level metrics and failure records to investigate bottlenecks and evaluate modifications under budget checks, experiment tracking, and human review. Together, these components provide a reusable foundation for extending the current object-removal pipeline to broader tasks such as joint audiovisual stylization.
Upon acceptance, we will publicly release our data engine to support further dataset construction and processing across audiovisual tasks.
\section{Limitations and Future Directions}
\label{sec:limitations}

\mypar{Interpreting LLM-based evaluation.}
Gemini 3.1 Pro~\citep{googledeepmind2026gemini31pro} assigns an AES of 11.10\% to the
Oracle--Oracle reference for the removal task.
The oracle reference consists of pseudo-ground-truth content
and does not guarantee complete target-sound removal.
Residual sounds, ambiguity between overlapping environmental
sources, and errors in the judge may contribute to the low score,
although their individual effects cannot be determined
from aggregate results.
We therefore complement LLM judgments with reconstruction,
perceptual, distributional, and synchronization metrics.
We also provide extensive video and audio examples on our
project page, allowing readers to directly assess target removal,
preservation of surrounding content, and audiovisual consistency.

\mypar{Imperfect pseudo-ground-truth supervision.}
Despite automated filtering and joint refinement,
pseudo-ground-truth targets may contain incomplete removals
or editing artifacts.
Reference-based metrics therefore measure agreement with
imperfect targets and may not fully capture editing quality.
Qualitative examples on our project page suggest that AVIO
can sometimes remove targets more completely than the
corresponding pseudo-targets while maintaining coherent
visual and acoustic content.
A possible explanation is that the pretrained text-to-audiovisual
model supplies generative priors that help refine imperfect
supervision rather than simply reproduce its artifacts.
This interpretation remains a hypothesis, and the observed
examples do not establish a systematic advantage over
the pseudo-ground-truth targets.

\mypar{Future data expansion.}
AVIOBench draws on existing public datasets and inherits
their limitations in recording quality, coverage, and diversity.
Our data engine provides a reusable workflow for extending
paired supervision to newer, higher-quality, and more diverse
audiovisual recordings.
Combining improved raw data with stronger editing and
verification models could improve target quality and broaden
the range of objects, sounds, and environments represented.
AVIOBench and the data engine thus provide a foundation
for expanding toward more diverse and controllable
audiovisual editing.

\end{document}